\documentclass[acmtog,authorversion]{acmart}

\AtBeginDocument{%
  \providecommand\BibTeX{{%
    \normalfont B\kern-0.5em{\scshape i\kern-0.25em b}\kern-0.8em\TeX}}}

\usepackage{graphicx}
\usepackage{amsmath}
\usepackage[]{algorithmic}
\usepackage{bm} 
\usepackage[ruled, vlined]{algorithm2e}
\usepackage{multirow}
\usepackage{xcolor}
\usepackage[shortcuts]{extdash}
\usepackage{caption}
\usepackage{subcaption}

\begin{document}

\setcopyright{acmcopyright}
\acmJournal{TOG}
\acmYear{2022} \acmVolume{1} \acmNumber{1} \acmArticle{1} \acmMonth{1} \acmPrice{15.00}\acmDOI{10.1145/3516521}

\title{HRBF-Fusion: Accurate 3D Reconstruction from RGB-D Data Using On-the-Fly Implicits}
\author{Yabin Xu}
\email{yabinxu007@gmail.com}
\orcid{0000-0002-1919-8673}
\affiliation{%
  \institution{Nanjing University of Aeronautics and Astronautics, China / Delft University of Technology}
  \country{The Netherlands}
}

\author{Liangliang Nan}
\affiliation{%
  \institution{Delft University of Technology}
  \country{The Netherlands}
}
\email{liangliang.nan@gmail.com}

\author{Laishui Zhou}
\affiliation{%
  \institution{Nanjing University of Aeronautics and Astronautics}
  \country{China}
  }
\email{zlsme@nuaa.edu.cn}

\author{Jun Wang}
\affiliation{%
  \institution{Nanjing University of Aeronautics and Astronautics}
  \country{China}
}
\email{wjun@nuaa.edu.cn}

\author{Charlie C.L. Wang}
\affiliation{%
  \institution{The University of Manchester, United Kingdom / Delft University of Technology}\country{The Netherlands}
}
\email{changling.wang@manchester.ac.uk}

\renewcommand{\shortauthors}{Xu et al.}

\begin{abstract}
Reconstruction of high-fidelity 3D objects or scenes is a fundamental research problem. Recent advances in RGB-D fusion have demonstrated the potential of producing 3D models from consumer-level RGB-D cameras. However, due to the discrete nature and limited resolution of their surface representations (e.g., point- or voxel-based), existing approaches suffer from the accumulation of errors in camera tracking and distortion in the reconstruction, which leads to an unsatisfactory 3D reconstruction. In this paper, we present a method using on-the-fly implicits of Hermite Radial Basis Functions (HRBFs) as a continuous surface representation for camera tracking in an existing RGB-D fusion framework. Furthermore, curvature estimation and confidence evaluation are coherently derived from the inherent surface properties of the on-the-fly HRBF implicits, which devote to a data fusion with better quality. We argue that our continuous but on-the-fly surface representation can effectively mitigate the impact of noise with its robustness and constrain the reconstruction with inherent surface smoothness when being compared with discrete representations. Experimental results on various real-world and synthetic datasets demonstrate that our HRBF-fusion outperforms the state-of-the-art approaches in terms of tracking robustness and reconstruction accuracy. 
\end{abstract}



\begin{teaserfigure}
\centering
\includegraphics[width=0.99\textwidth]{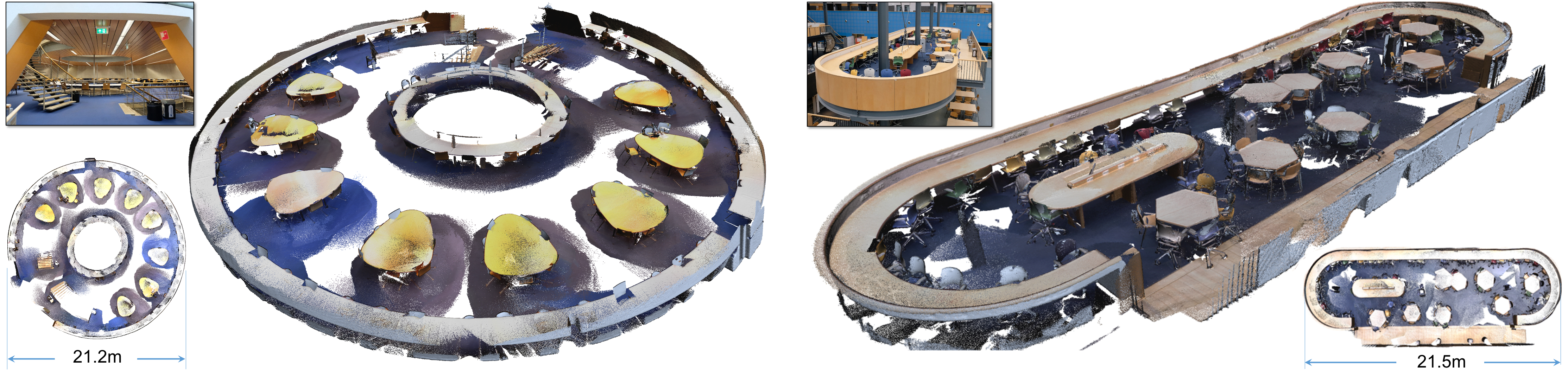}\vspace{-5pt}
\caption{Reconstruction of two large indoor scenes: (left) a study room of a university library and (right) a study platform in a grand hall of an academic building. The original two sequences consists of 16,128 (library) and 10,930 (study platform) RGB-D image frames and the reconstructed model consists of 7,488,867 (library) and 7,904,727 (study platform) points respectively. The average processing speed of our approach is around 43ms per frame, which demonstrates a nearly real-time performance. RGB-D data in these two experiments are captured by a Microsoft Kinect v1 sensor with a resolution of $640 \times 480$. Progressive results of the reconstruction can be found in the supplementary video.
\vspace{10pt}
}
\label{figTeaser}
\end{teaserfigure}

\maketitle

\section{Introduction}
\label{secIntroduction}
Reconstruction of high-fidelity 3D objects or scenes is vital to applications such as augmented~/~virtual reality, digital fabrication, and robotics. With the increasing popularity of consumer-level depth cameras (e.g., Microsoft Kinect), 3D information, in the form of RGB-D images or point clouds, can be easily obtained. A lot of reconstruction systems targeting on producing surface models of small-scale objects or large scenes~\cite{Keller2013,lefloch2017,Whelan2016ElasticFusionRD,Dai2017,Cao2018,zhou2015,choi2015robust} have been introduced since the pioneering work of \textit{KinectFusion}~\cite{Newcombe2011}. Despite the advances in 3D reconstruction in the last decade, obtaining high-quality 3D models from consumer-grade depth cameras remains an open problem due to the following two main issues.
\begin{itemize}
\item \textit{Imperfect Surface Representation}: Existing approaches lack an accurate surface representation that facilitates high-fidelity reconstruction while being memory efficient and computationally affordable. The volumetric representation is widely used for RGB-D reconstruction systems~\cite{Dai2017,Chen2013, NieBner2013} following \textit{KinectFusion}~\cite{Newcombe2011}. However, a commonly used implementation with fixed-size resolution lacks adaptiveness~\cite{Dai2017,Chen2013, NieBner2013}, which tends to generate over-smoothed surfaces in the regions with geometric details. The alternative surface representation~\cite{Keller2013} -- surfel, which predicts geometry by ray-to-plane surfel splatting, works poorly in high-curvature regions and is also prone to failure due to noises and outliers.
    
\item \textit{Camera Tracking Error}: Imprecision registration based on Iterative Closest Point (ICP) or its variants~\cite{Besl1992, Rusinkiewicz2001} is usually applied for camera pose estimation between RGB-D frames, where distortion errors are accumulated and can become significant in featureless regions. Research efforts have been paid to resolve the problem through global optimization~\cite{choi2015robust, Zhou:2013} or additional information provided by the RGB-D camera (e.g., geometric~\cite{zhou2015, lefloch2017} and photometric~\cite{Whelan2016ElasticFusionRD} information), to derive a weighted variant of the ICP scheme to reduce camera tracking drift. 
In recent pipelines (e.g.,~\cite{Whelan2016ElasticFusionRD,Dai2017,Cao2018}), both strategies are applied to improve the result of reconstruction. 
\end{itemize}
The issue of camera tracking is also suffered from the lack of good surface representation when geometric cues are employed to enhance ICP registration.

\subsection{Our method}
To address the aforementioned issues, we propose HRBF-Fusion, a new method using on-the-fly HRBF implicits for high-accurate camera tracking and high-fidelity 3D reconstruction. The core of our method is a voxel-free implicit surface representation, i.e., the closed-form HRBF surface approximation that gracefully benefits multiple key stages of the reconstruction pipeline, including preprocessing, camera pose estimation, and depth map fusion. The 3D reconstruction pipeline used in our tests is a variant of \textit{ElasticFusion}~\cite{Whelan2016ElasticFusionRD} and \textit{ORB-SLAM2}~\cite{Artal2017}, in which the tracking-and-fusion steps of ElasticFusion 
are used to generate submaps and the ORB-based local-to-global optimization routine is used to obtain a global consistent 3D model for large scenes. In contrast, we evaluate both the global model and the new RGB-D frames as continuous but compactly-supported HRBF surfaces to produce robust curvature estimation and reconstruction-indicated confidence maps. With the help of these HRBF surfaces, more reliable camera tracking and depth map fusion can be achieved. In summary, we make the following contributions:
\vspace{-5pt}
\begin{itemize}
\item A method to evaluate a continuous surface effectively and efficiently on both the global model and the acquired RGB-D frame by using on-the-fly HRBF implicits;

\item A robust and efficient curvature evaluation method based on the on-the-fly HRBF implicits, leading to a dramatic improvement in camera tracking based on the curvature-weighted registration;

\item A reconstruction-indicated confidence evaluation method, also based on efficient HRBF surface evaluation, can significantly reduce the impact of noises and outliers in both camera tracking and depth-image fusion. 
\end{itemize}
\vspace{-5pt}
As a consequence, we develop a more robust reconstruction system for high-fidelity online surface reconstruction, which also shows good scalability to large scenes.

\subsection{Related work}
\subsubsection{Geometric representation} 3D reconstruction within a commodity RGB-D camera has been extensively studied in the past decade. A key ingredient toward a high-quality 3D reconstruction system is the underlying representation for camera pose estimation and depth map fusion. Different representations have been proposed, including volumetric representation~\cite{Curless:1996, Newcombe2011, NieBner2013, zhang2017cufusion, Dai2017}, surfel-based representation~\cite{weise2009, Keller2013, Whelan2016ElasticFusionRD, Cao2018}, height field~\cite{Meilland2013}, probability-based representation~\cite{Dong_2018_ECCV}, and 2.5D depth map~\cite{Gallup20103D}. A recently trend is to solve the problem by using neural implict representation for shape generation~\cite{liu2020neural, Huang2021DI,Sucar2020NodeSLAM,Huang2021Real,Sucar_2021_ICCV} and using learing-based method for depth fusion~\cite{Weder2020CVPR,Weder2021CVPR,Bozic2021NIPS}. Here we provide a compact solution by using a closed-form representation for the on-the-fly implicts.

Following the pioneering work of \textit{KinectFusion}~\cite{Newcombe2011} that applied a Truncated Signed Distance Field (TSDF) \cite{Curless:1996} for modeling integration, volumetric representation has demonstrated promising results for reconstructing small-scale scenes. Because of its implementation on GPU for real-time tracking and fusion, volumetric representation becomes more and more popular \cite{Chen2013,NieBner2013,Dai2017,MeeritsCVM2018}. The original uniform-grid KinectFusion has a fundamental limitation (i.e., the lack of scalability), which leads to expensive memory consumption for reconstructing fine details. Recently, a learning-based TSDF was adopted to represent the geometry under reconstruction \cite{Sun_2021_CVPR}. Although methods have been developed to alleviate this by exploiting sparsity in the TSDF representation~\cite{Chen2013, NieBner2013}, the quality of local reconstruction still depends on the resolution to partition the space which is related to the scale of the scene. 

Kelly et al.~\shortcite{Keller2013} proposed a surfel-based representation method to solve the scalability issue and has presented comparable results against volumetric methods on flat or smooth regions. In their method, a ray-to-plane surfel rendering algorithm is used to predict the model for real-time camera tracking. The method has been applied to real-time reconstruction systems~\cite{Whelan2016ElasticFusionRD, Cao2018, lefloch2017}. However, the linear ray-to-plane based shape prediction is sensitive to noises in particular on the high-curvature surface regions. Hence reconstructed models are often distorted when there are noises in high-curvature regions. Implicit moving least-squares (IMLS) surface was employed in \cite{Liu_2021_CVPR} to achieve a better shape representation in their learning-based 3D reconstruction. However, the evaluation of IMLS is less efficient. Differently, we predict the shape from surfels by using closed-form HRBF implicits which makes our system is memory-efficient and robust.

Radial Basis Functions was employed in \cite{Car2001RBF} for surface reconstruction. In this method, the computation is however very time-consuming, and it also requires the provision of auxiliary `off-surface' points. Liu et al.~\shortcite{liu2016} introduced closed-form HRBF implicits using quasi-interpolation, which has demonstrated its capability of generating surface reconstruction in high quality and high efficiency. Inspired by this work, we explore the possibility to incorporate the closed-form HRBF implicits with the inherent surface properties for noise-resistant camera tracking and high-quality 3D reconstruction. It is also worthy to notice that Schöps et al.~\shortcite{Schops2019PAMI} recently developed an online mesh construction method for reconstruction refinement; nevertheless, camera poses are required as additional input for their method. Differently, our on-the-fly HRBF implicits are directly devoted to camera tracking and RGB-D reconstruction.
%

\subsubsection{Camera tracking} An important issue in the real-time RGB-D surface reconstruction system is the drift of camera tracking caused by the instability of the frame-to-model registration.

One of the reasons that cause the instability of registration is the presence of noise and outliers. To mitigate the impact of noise and outliers, Jian and Vemuri~\shortcite{jian2010robust} proposed to use the Gaussian Mixture Model (GMM) to describe the distribution of both template and point set. Not only geometric but also color information has been conducted for probabilistic registration~\cite{danelljan2016probabilistic}. Although robust, these probabilistic registration approaches are time-consuming which makes them ineligible for real-time reconstruction from a sequence of input RGB-D frames. Others tend to evaluate the reliability of an input raw depth map by analyzing the inherent property of depth cameras (e.g.,~\cite{reynolds2011capturing}). Similarly, a distortion-based model is employed in~\cite{Keller2013} which weights measurements based on the assumption that the depth data captured near the center of a sensor are more accurate. Recently, a voting mechanism is introduced in~\cite{Cao2018} to evaluate the confidence of depth map for generalized ICP~\cite{segal2009generalized} by using the time-coherence between nearby frames. In this paper, we propose a novel reconstruction-indicated confidence metric to exploit the underlying uncertainty on each depth map.

Another reason for tracking drift is the lack of salient geometric features in the scene which leads to slippery registration. As depth cameras are commonly equipped with an additional RGB camera, colors are used as additional information to form a joint optimization problem~\cite{Whelan2016ElasticFusionRD, Godin1994} or to pre-align the depth map with color-based features~\cite{Henry2012}. Yang et al.~\shortcite{ShengCGF2017} incorporated visual saliency into a volumetric fusion pipeline to achieve high-quality object reconstruction. Other geometric features have also been considered in other approaches to add weights in the optimization for registration, including contour cues~\cite{zhou2015}, planar structures, and repeated objects~\cite{Zhang2015}, patch co-planarity~\cite{Shi2018} and curvatures~\cite{lefloch2017}. Among them, the curvature is very general and can be evaluated in all regions. Several methods of curvature estimation have been discussed in~\cite{lefloch2017}, among which the method of adjacent-normal cubic approximation~\cite{Goldfeather2004} is concluded as the most robust curvature estimator. However, this method needs to solve a $7 \times 7$ linear system at every point, which hinders its usage for real-time applications even with the implementation on GPU (ref.~\cite{lefloch2017}). By the requirement of real-time performance, Lefloch et al.~\shortcite{lefloch2017} selected the chord-and-normal-vectors (CAN) approach~\cite{ZhangCurvatureEstimation2008} for curvature estimation.
%
%
The camera drift can be reduced by integrating curvature information into the ICP framework with higher weights in high curvature regions~\cite{lefloch2017}. However, the curvature evaluation in their approach is not robust when input RGB-D data becomes noisy. This is crucial as the input frames from consumer-level RGB-D cameras are often contaminated with noises and outliers. In our approach, we use curvature as additional information in both tracking and fusion stages -- but differently, curvature in our approach is robustly extracted from the continuous surfaces represented by HRBF implicit.  

\subsubsection{Accumulated error} Apart from focusing on the error sourced from the frame-by-frame registration, methods have been developed to correct the error accumulation in camera pose estimation and global 3D model in both online \cite{whelan2012kintinuous,Wasenmuller2016,Dai2017,Cao2018,WangCVM2017} and offline~\cite{choi2015robust,Li20133DS,Zhou:2013,zhou2013elastic} mode, where the offline methods are  time-consuming. 

For online correction,~Whelan et al.~\shortcite{Whelan2016ElasticFusionRD} proposed a system that divides the reconstructed model into active (recently captured frames) and inactive parts. When the registration between active and inactive parts is successful, an optimization-based deformation is applied to deform the active part to fuse into the inactive part. However, the routine does not provide a way to fix the errors that have already been inherited into the inactive part. Yang et al.~\shortcite{ShengTOG2020} proposed a noise-resilient panoramic scanning approach that uses robot-mounted multiple RGB-D cameras to obtain high-quality 3D models of the scene. A different strategy is applied in the area of simultaneous localization and mapping (SLAM)~\cite{klein2007parallel, engel2013semi, forster2014svo, mur2015orb, Artal2017}, where drift-free pose estimation has been extensively studied. The basic idea of these approaches is to minimize the reprojection
error across frames or distribute camera pose estimation error across the pose-graph constructed by the co-visibility between frames. While focusing on different problems, these approaches do not provide a method to correct dense 3D models generated from depth map fusion. To solve this problem, submap-based online reconstruction systems (e.g., \cite{Dai2017,Cao2018}) are proposed to correct the camera poses and minimize the geometric error of 3D models in an integrated manner. In our system, we adopt a similar submap-based hierarchical optimization for the steps of close-loop detection, camera pose, and 3D model correction.

\section{Overview}
\label{secOverview}

\begin{figure}[t]
\centering\vspace{5pt}
\includegraphics[width=0.95\linewidth]{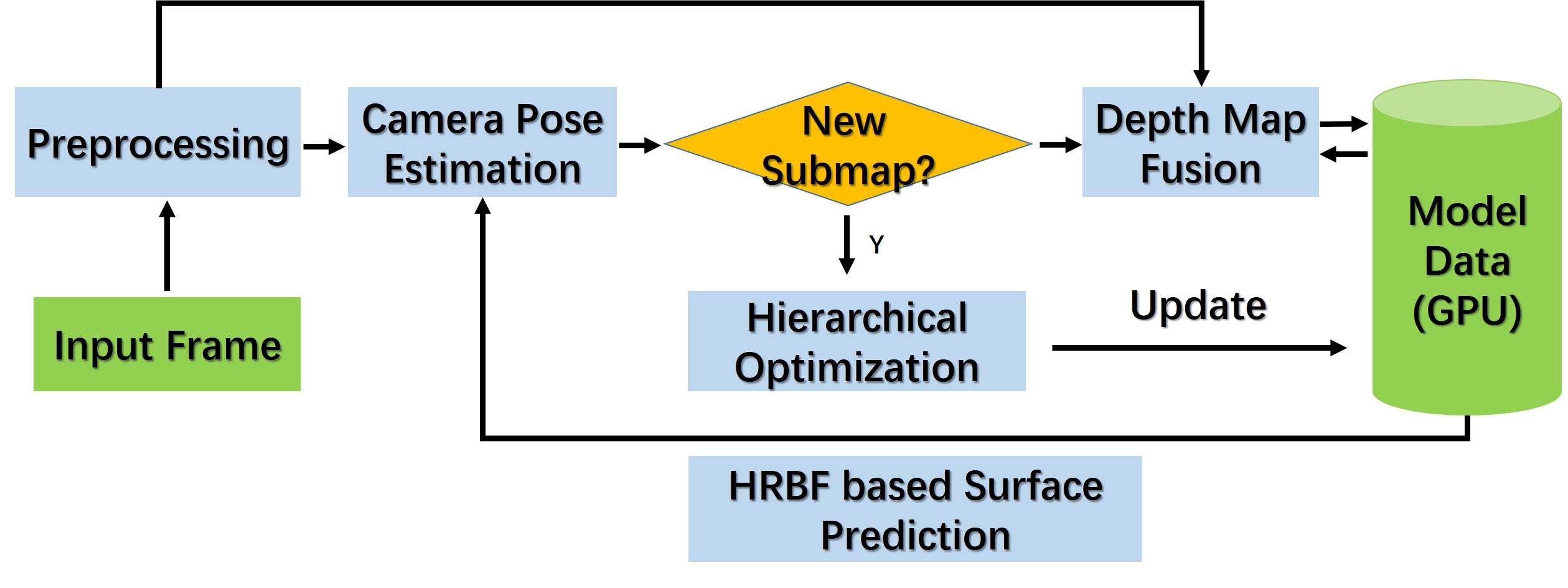}
\vspace{-5pt}
\caption{Framework of the proposed RGB-D reconstruction system.}\label{overview}\vspace{-10pt}
\end{figure}

We utilize closed-form HRBF implicits for on-the-fly surface evaluation for the global model, which replaces the commonly used discrete surface representations of existing reconstruction systems and plays a vital role in the key stages of the pipeline to improve tracking robustness and reconstruction accuracy. 
We adopt a computational framework similar to prior systems~\cite{Newcombe2011, NieBner2013, Keller2013, Dai2017, lefloch2017} for reconstruction (see Fig.~\ref{overview}). The functionality of closed-form HRBF implicits is utilized in various stages of the framework.

The global model $\mathbf{M}$ is represented by a set of unorganized points where each point is associated with attributes\footnote{Variables evaluated on the global model are represented by symbols with `$\bar{\; }$' head throughout the paper.} including its position $\bar{\mathbf{v}} \in \mathbb{R}^3$, normal $\bar{\mathbf{n}} \in \mathbb{R}^3$, support size $\bar{r} \in \mathbb{R}$, confidence value $\bar{c} \in \mathbb{R}$, and two principal curvature values $\bar{\kappa}_{1}$, $\bar{\kappa}_{2} \in \mathbb{R}$. This is a highly scalable representation, which can be considered as an enriched surfel representation~\cite{Keller2013,Cao2018}.

When capturing a new RGB-D frame $\mathbf{F}=\{\mathbf{D}, \mathbf{C}\}$ with $\mathbf{D}$ and $\mathbf{C}$ denoting the depth map and the color map respectively, the RGB-D frame $\mathbf{F}$ is fused into the global model by applying the following key steps:
\begin{itemize}
\item \textit{Preprocessing}: Continuous surfaces are evaluated in the input RGB-D frame and on the global model by using the on-the-fly HRBF implicits respectively (Section \ref{subsecSurfaceEvaluation}). Note that the HRBF surface for the global model is evaluated in the previous frame of the scanning sequence. With the help of robust HRBF surface evaluation, a curvature map (Section \ref{subsecCurvature}) and a reconstruction-indicated confidence map (Section \ref{subsecConfidenceMap}) are evaluated in the input frame to enhance the robustness of our reconstruction pipeline. 

\item \textit{Camera pose estimation}: The purpose of this step is to obtain the transformation between the input frame and the current global model. We adopt a variant ICP algorithm based on the point-to-plane metric with specially designed searching and weighting schemes to align it to the surface predicted from its last pose. Unlike existing RGB-D reconstruction systems based on discrete surface representations, our accurate and robust local surface reconstruction based on HRBF implicits improves the robustness in both the correspondence search (Section \ref{subsecCorrespondSearch}) and the optimization of registration (Section \ref{subsecRegistration}). On-the-fly calculated curvatures and normals are stored in local but `dense' maps for camera pose estimation, which can avoid the problem caused by sparsity in a global map.

\item \textit{Depth map fusion}: To integrate a new frame into the global model with a valid pose, correspondences between vertices of the input frame and the points in the global model are established based on an index map that is obtained by rendering the index of each model point into a texture~\cite{Keller2013}. After that, the input vertices with their attributes are merged into the global model using a confidence-weighted average (Section \ref{secDepthFusion}). Similar to other surfel-based approaches (e.g.,~\cite{Keller2013,Cao2018}), attributes stored on the global model are employed to conduct the fusion.
\end{itemize}
These steps are repeated until the relative translation between the first frame and the current frame exceeds a certain threshold. Then, the global model formed by already registered and fused frames will be treated as a submap. 

With reliable geometric and photometric enhanced registration, high-quality camera tracking and surface reconstruction can be achieved for relatively small objects. When reconstructing large scenes by long-range scanning, a local-to-global optimization scheme similar to \cite{Cao2018} is applied between submaps to further alleviate the accumulation of errors in camera tracking by using the ORB features~\cite{Rublee2012ORB}.


\section{Geometric Cues by HRBF Implicits}
\label{secGeometryCues}
In this section, we first introduce the method of surface prediction with closed-form HRBF implicits. After that, the robust curvatures and the reconstruction-indicated confidence map can be generated from the on-the-fly HRBF surfaces. 

HRBF implicits have been used to reconstruct an implicit function from scattered Hermite points \cite{Macedo2011HermiteRB}. Given a point set $\textbf{P} =\{\mathbf{p}_1, \mathbf{p}_2,...\mathbf{p}_n\}$ with corresponding normals $\textbf{N} =\{\mathbf{n}_1, \mathbf{n}_2,...\mathbf{n}_n\}$, a function $f$ interpolating the positions and the normals can be defined as
\begin{equation} \label{eqn:hrbf_implicits}
f(\mathbf{x}) = \sum_{j = 1}^{n}\{\alpha_j\psi (\mathbf{x}-\mathbf{p}_{j})-<\boldsymbol{\beta}_{j},\mathbf{\triangledown} \psi(\mathbf{x}-\mathbf{p}_{j})>\},
\end{equation}
where $\langle \cdot, \cdot \rangle$ denotes the dot-product of two vectors, and $\mathbf{\triangledown}$ is the gradient operator. The \textit{Compactly Supported Radial Basis Functions} (CSRBF) \cite{Wendland1995} are applied as the kernels because of their numerical stability and the on-the-fly nature. Specifically, we have
\begin{equation} \label{eqn:kernel_function}
\psi(\mathbf{x}-\mathbf{p}_{j}) = \begin{cases}
           (1- \frac{d}{r} )^4 (\frac{4d}{r}  + 1 ), & d \in [0, r],\\
           0,             &otherwise, 
           \end{cases}
\end{equation}
where $d=\| \mathbf{x}-\mathbf{p}_{j} \|$ is the Euclidean distance between the query point and the corresponding CSRBF kernel and $r$ is the support size. The coefficients $\alpha_j \in \mathbb{R}$ and $\boldsymbol{\beta}_j \in \mathbb{R}^3$ can be computed from the following constraints: $f(\mathbf{p}_i) = 0$ and $\mathbf{\triangledown} f(\mathbf{p}_i) = \mathbf{n}_i$ on all given points $\mathbf{p}_{i=1,\ldots,n}$. Instead of solving a $4n \times 4n$ linear system, a closed-form function was proposed in \cite{liu2016} to approximate the HRBF implicits as 
\begin{equation} \label{eqn:hrbf_approximation_solution}
    \hat{f}(\mathbf{x}) = - \sum_{j = 1}^{n}<\frac{r_j^2}{20 + \eta r_j^2}\mathbf{n}_j, \triangledown\psi(\mathbf{x} - \mathbf{p}_j)>,
\end{equation}
where $r_j$ is the support size of kernel centered at $\mathbf{p}_j$. The value of $r_j$ should be determined to cover at least 8 neighboring kernels for constructing a locally continuous surface~\cite{liu2016} around $\mathbf{p}_j$. $\eta = 1.0 \times 10^6$ is employed as the regularization coefficient for points evaluated in the unit of meter~\cite{liu2016}.  With such a closed-form surface representation, solving the linear system can be avoided. This enables a method for efficient and on-the-fly surface evaluation, which is very important for real-time reconstruction.

\begin{figure}[t]
	\setlength{\unitlength}{0.1\textwidth}
	\begin{picture}(5, 1.25)
	\put(0.05,-0.05){\includegraphics[width=\linewidth]{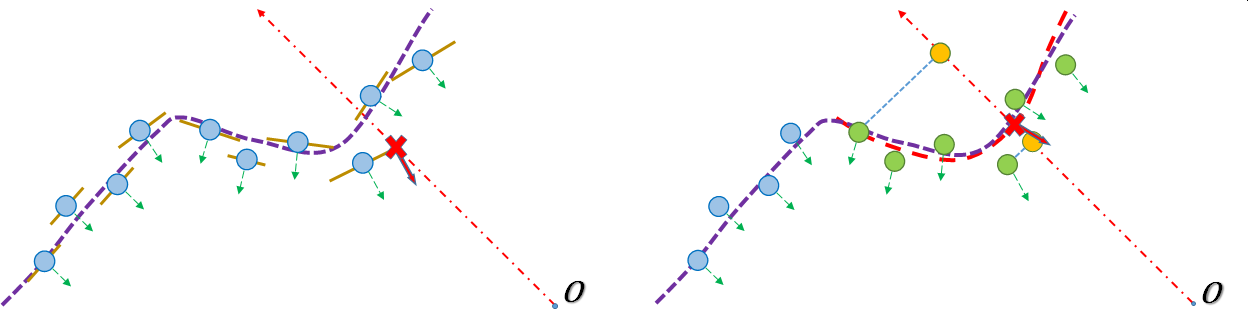}}
	\put(0.7,0){\scriptsize Surfel splatting}
	\put(3.3,0){\scriptsize Our approach}
	\put(3.7,1.0){\scriptsize $\mathbf{p}_f$}
	\put(3.67,0.725){\scriptsize $\mathbf{p}_m$}
	\put(4.05,0.53){\scriptsize $\mathbf{p}_n$}
	\end{picture}
	\caption{An illustration of the surfel splatting (left)~\cite{Keller2013,Cao2018} and our HRBF-based (right) surface prediction methods. The red cross in each figure represents the intersection between the ray (red dashed line) and the global model points (blue dots).}
	\label{fig:illustration_hrbf_prediction}
\end{figure}
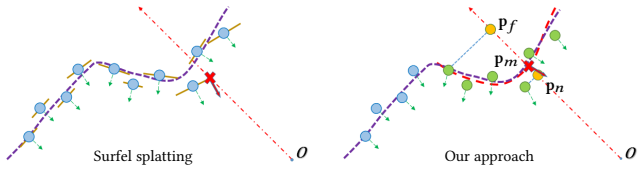

\subsection{Surface evaluation}\label{subsecSurfaceEvaluation}
In our approach, continuous surfaces are evaluated for both the newly captured depth image and the global model by using the on-the-fly HRBF implicits. Specifically, two surfaces are evaluated on all pixels of two frames -- the input RGB-D frame for a local model and the previous frame in the scanning sequence for the global model. Similar to the raycasting method of~\cite{Newcombe2011}, we predict the surface points for a pixel $\mathbf{u}$ at the current pose by intersecting the HRBF local surface with the ray from the camera optical center to the corresponding point in the image plane (see Fig.~\ref{fig:illustration_hrbf_prediction}). In contrast to the popular surfel-based surface prediction method~\cite{Cao2018,Keller2013} that searches for the nearest (from the viewpoint) discrete point within a radius (see the left of Fig.~\ref{fig:illustration_hrbf_prediction} for illustration), our method takes advantage of the smooth nature of the surface and thus is more robust to noise and outliers. 

For the surface evaluation in a frame by HRBF implicits, we choose the kernels that are closer to the viewpoint while discarding kernels that have greater depth deviation from the nearest model point due to depth discrepancy. After obtaining a local set of kernels that define the HRBF surface on a viewing ray, we project the kernels' centers onto the ray to form a searching interval $[\mathbf{p}_n, \mathbf{p}_f]$, where $\mathbf{p}_n$ is the nearest point and $\mathbf{p}_f$ is the furthest one along the viewing ray. The model point $\mathbf{p}_m$ is supposed to lie in the interval to satisfy $\hat{f}(\mathbf{p}_m)=0$, which can be obtained by a binary searching algorithm (see the right of Fig.~\ref{fig:illustration_hrbf_prediction} for an illustration). After determining the position of a surface point, other attributes at $\mathbf{p}_m$ such as colors can be predicted from its nearest kernel. Note that, this ray-intersection based surface evaluation can run in highly parallel mode on the many cores of GPU. Specifically, we implement the surface evaluation of HRBF implicits in a fragment shader that is used for per-pixel operation with each viewing ray defined on pixels. The input vertex map and normal map are bound with the fragment shader for local searching. The HRBF implicits are constructed and evaluated within the fragment shader. The outputs are the texture maps bound with a frame buffer, which are the predicted surface points and their corresponding attributes. 

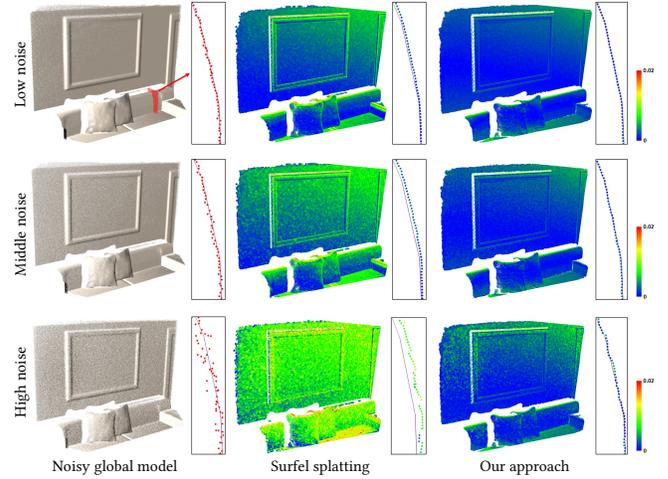
\begin{figure}
	\setlength{\unitlength}{0.1\textwidth}
	\begin{picture}(5,3.55)
	\put(0.1,0.1){\includegraphics[width=0.98\linewidth]{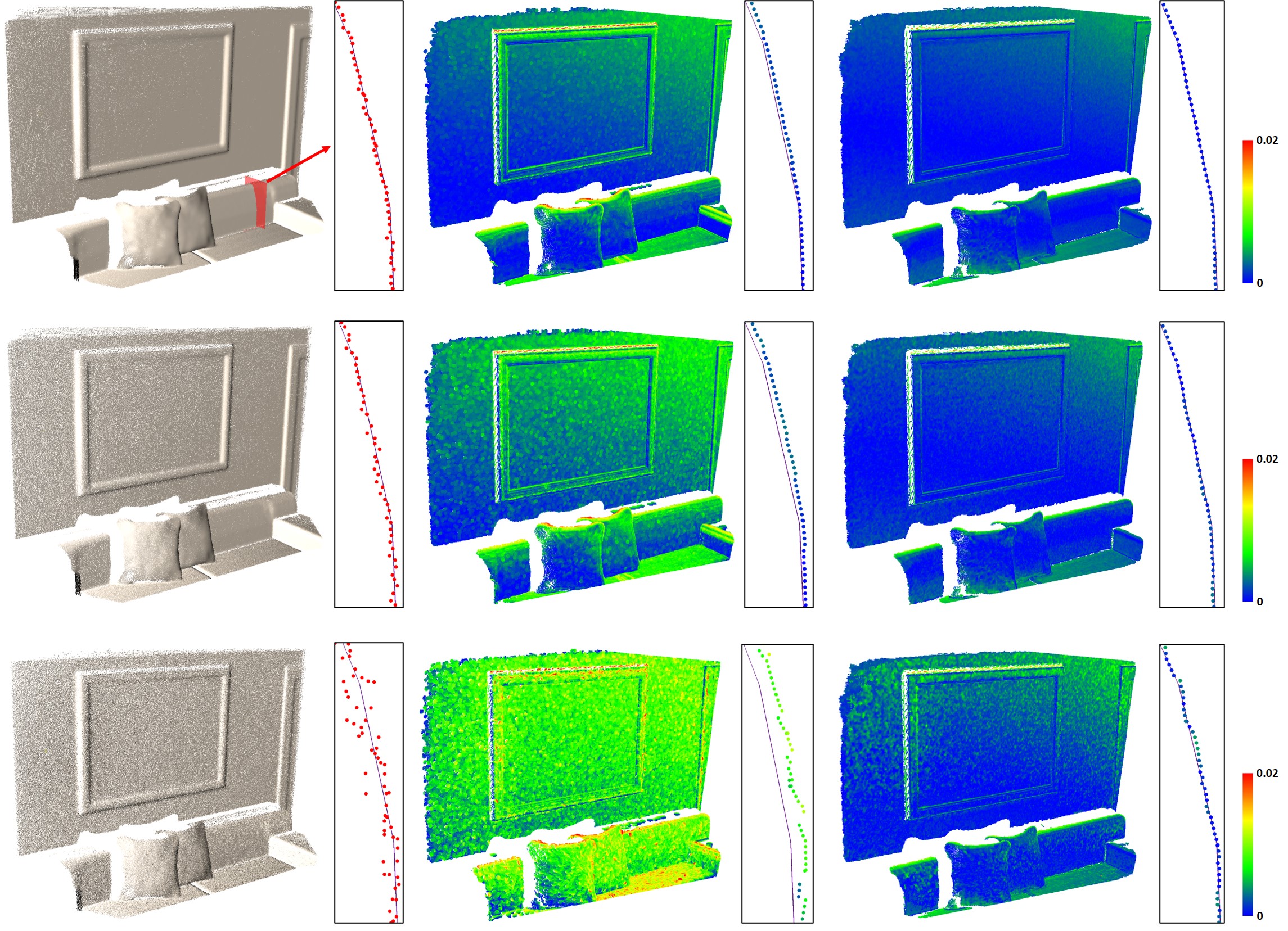}}
	\put(0.28,0.0){\scriptsize Noisy global model}
	\put(1.9,0.0){\scriptsize Surfel splatting}
	\put(3.45,0.0){\scriptsize Our approach}
	\put(0.0, 2.7){\scriptsize \rotatebox{90}{Low noise}}
	\put(0.0, 1.45){\scriptsize \rotatebox{90}{Middle noise}}
	\put(0.0, 0.4){\scriptsize \rotatebox{90}{High noise}}
	\end{picture}
\caption{Comparison of the predicted vertex map generated by surfel splatting (middle column) and HRBF local surface reconstruction (right column) on the same global model (left column). The test is conducted on the \textit{lr kt1} example from \textit{ICL-NUIM}~\cite{handa:etal:ICRA2014} by adding three levels of Gaussian noise with the standard deviations as $\sigma=3.0$, $\sigma=6.0$ and $\sigma=12.0$ respectively, where the ground truth of the geometry and camera poses are provided. Note that the global models are generated by fusing multiple RGB-D frames (i.e. 1-68) using the same strategy of \cite{Keller2013} and the ground-truth camera poses. Colors indicate the unsigned distances from the points to the ground-truth 3D model.}
\label{fig:accurate_representation_distance_visulization}
\end{figure}

\begin{figure}[t]
	\setlength{\unitlength}{0.1\textwidth}
	\begin{picture}(5.25,2.35)
	\put(0.2,0.04){\includegraphics[width=0.95\linewidth]{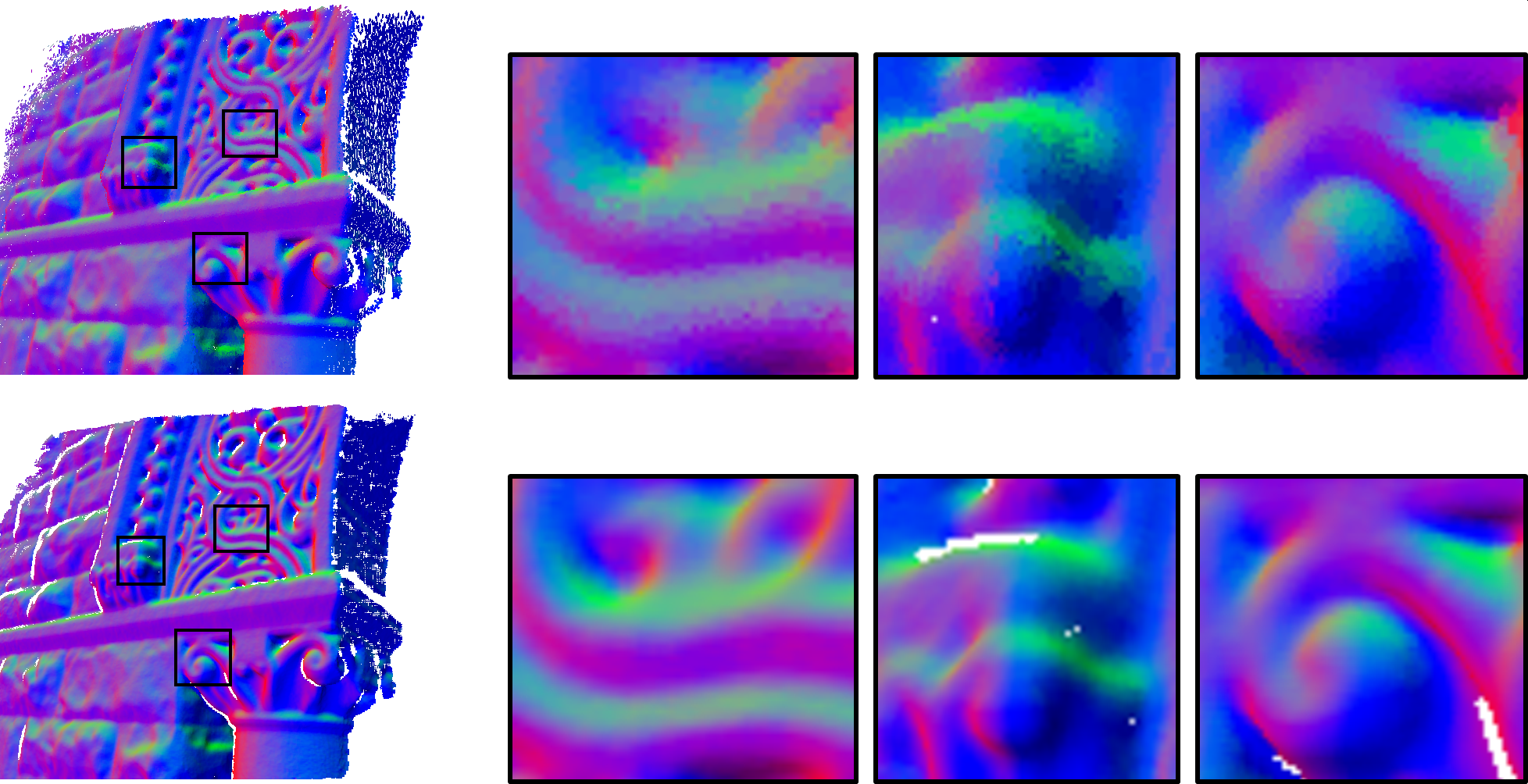}}
	\put(0.02, 0.2){\scriptsize \rotatebox{90}{Our approach}}
	\put(0.02, 1.4){\scriptsize \rotatebox{90}{Surfel splatting}}
	\end{picture}
	\caption{Comparison of the normal maps generated by the ray-to-plane surfel splatting method (top row) and our HRBF-based prediction method (bottom row) on 
		the \textit{stone wall} from 3D Scene Data~\cite{zhou2015}.
	}\label{fig:accurateNormalrepresentation}
\end{figure}

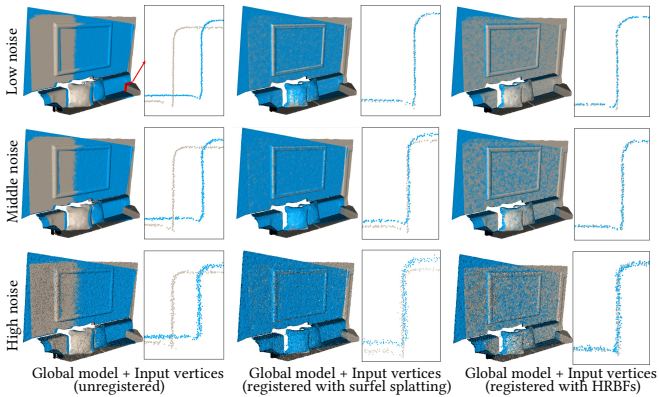
\begin{figure}
	\setlength{\unitlength}{0.1\textwidth}
	\begin{picture}(5,3.0)
	\put(0.1,0.2){\includegraphics[width=0.98\linewidth]{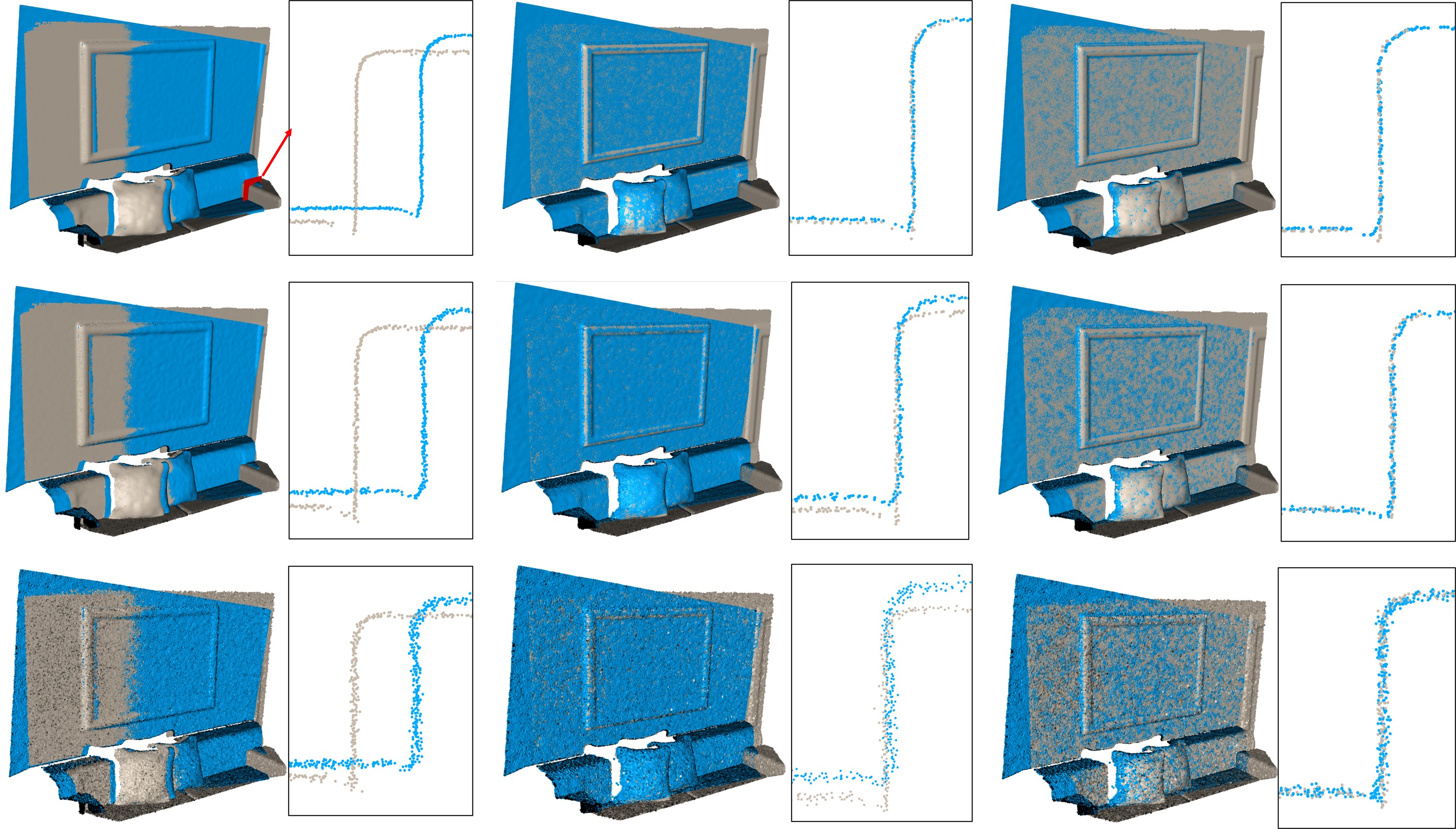}}
	\put(0.19,0.1){\scriptsize Global model + Input vertices}
	\put(0.50,-0.0){\scriptsize (unregistered)}
	\put(1.8,0.1){\scriptsize Global model + Input vertices}
	\put(1.75,-0.0){\scriptsize (registered with surfel splatting)}
	\put(3.4,0.1){\scriptsize Global model + Input vertices}
	\put(3.55,0.0){\scriptsize (registered with HRBFs)}
	\put(0.0, 2.2){\scriptsize \rotatebox{90}{Low noise}}
	\put(0.0, 1.25){\scriptsize \rotatebox{90}{Middle noise}}
	\put(0.0, 0.3){\scriptsize \rotatebox{90}{High noise }}
	\end{picture}
\caption{Comparison of registration between the global model and the input vertices under different noise levels. Similar to already discussed in Fig.~\ref{fig:accurate_representation_distance_visulization}, the global model is obtained by fusing multiple RGB-D frames (i.e. 1-68) using the same strategy of \cite{Keller2013} and the ground-truth camera poses. The left column shows the initial alignment of the global model and the input vertices, whereas the other two columns show registration results based on surfel splatting (the middle column) and our HRBF based method (the right column).}
\label{fig:Comparison_accurate_representation_registration}
\end{figure}

For surface evaluation on a global model $\mathbf{M}$, the stored points $\{\bar{\mathbf{v}} \}$ will be used as the kernels of HRBF implicits. The resultant intersection points are stored in a 3D vertex map $\bar{\mathbf{V}}$. The normal at each intersection point $\mathbf{p}_m$ can also be obtained from the gradient as $\triangledown \hat{f}(\mathbf{p}_m) / \|\triangledown \hat{f}(\mathbf{p}_m) \|$. The resultant normal map is denoted by $\bar{\mathbf{N}}$. With the help of the closed-form HRBF implicits, we are able to predict $\bar{\mathbf{V}}$ more accurately -- see the comparison with surfel splatting on a model with ground-truth geometry (Fig.~\ref{fig:accurate_representation_distance_visulization}).
%
The experiment is conducted on the \textit{lr kt1} example from the synthetic dataset \textit{ICL-NUIM}~\cite{handa:etal:ICRA2014} with the ground-truth geometry and camera poses provided. To evaluate the sensitivity to noise, the input RGB-D frames are contaminated by adding different levels of Gaussian noise. The global models are obtained by fusing multiple (i.e, 1-68) input frames with the ground-truth camera poses, while the same strategy of \cite{Keller2013} is adopted for depth map fusion. As can be observed from the cross-sectional views in Fig.~\ref{fig:accurate_representation_distance_visulization}, the increased level of noise makes the points in the global model corrupt gradually. The surfel splatting method results in imprecise prediction of the underlying surface when highly noisy input is given. In contrast, the vertex map predicted by our method can properly represent the underlying surface.
Moreover, smoother normal maps can be generated by our method (Fig.~\ref{fig:accurateNormalrepresentation}). 
Note that an accurate and robust prediction of geometry is the key ingredient to the high accuracy in camera pose estimation (Section~\ref{secCameraPose}). With our HRBF-based local surface reconstruction, the accuracy of geometry prediction and thus the registration is dramatically improved (see Fig.~\ref{fig:Comparison_accurate_representation_registration} for an example).

The kernels for surface evaluation in the input RGB-D frame are determined differently. Preprocessing is needed before applying the HRBF based surface evaluation. Given an input frame with the depth map and the color map, its corresponding 3D vertex map $\mathbf{V}$ is computed using the camera intrinsic matrix $\mathbf{K}$ by following the same steps as KinectFusion~\cite{Newcombe2011}. After applying a bilateral filter to reduce noise while preserving discontinuity in the depth map $\mathbf{D}$, the corresponding 3D vertex for each pixel $\mathbf{u} = (x, y)^T \in \mathbb{R}^2$ is computed as $\mathbf{V}(\mathbf{u}) = \mathbf{D}(\mathbf{u})\mathbf{K}^{-1}(\mathbf{u}^\top, 1.0)^T$. The corresponding normal map $\mathbf{N}$ can be derived from $\mathbf{V}$ by central difference. Besides, we assign each vertex with a support size $\mathbf{S}(\mathbf{u})$ for local HRBF surface evaluation. To construct a continuous HRBF surface, the support size of a kernel should cover at least $k$ other kernels (i.e., $k=8$ according to~\cite{liu2016}). The $k$-nearest neighbor for each pixel $\mathbf{u}$ is first obtained by searching the vicinity of $\mathbf{u}$ in a window patch (i.e., $7\times7$) of the filtered vertex map $\mathbf{D}$. The support size is assigned as the distance between a kernel and its $k$-th nearest neighbor. Lastly, the ray-intersection based surface evaluation is conducted in the input RGB-D frame to update its vertex map $\mathbf{D}$ and normal map $\mathbf{N}$.


\begin{figure}[t]
	\setlength{\unitlength}{0.1\textwidth}
	\begin{picture}(5.25,3.2)
	\put(0,0.04){\includegraphics[width=0.47\textwidth]{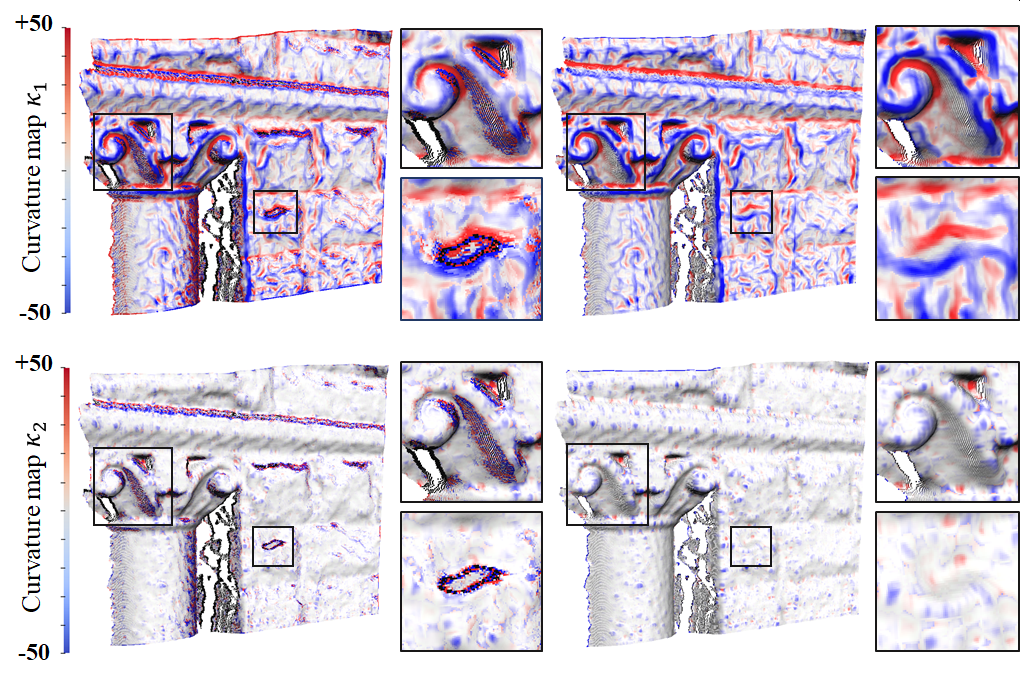}}
	\put(0.9, 0){\scriptsize \cite{lefloch2017}}
	\put(3.4, 0){\scriptsize Our approach}
	\end{picture}
\caption{Comparison of principal curvature estimated by \cite{lefloch2017} versus our method. The black points indicate the corresponding curvature values are out of a range of $[-300, 300]$. Note that, $|\kappa_{1}| , |\kappa_{2}| >300$ means the radius of curvature is already less than $3 mm$. These are geometric details that cannot be captured by RGB-D cameras -- i.e., unreliable estimation.}\label{fig:comparison_curvature_estimation}
\end{figure}

\subsection{Robust curvature}\label{subsecCurvature}
The principal curvature map $\mathbf{\kappa}$ is evaluated by the on-the-fly HRBF implicits in an input RGB-D frame, which provides important clues in the registration step (Section \ref{subsecRegistration}). Benefit from the continuous surface representation provided by HRBF implicit, the mean curvature $H$ and the Gaussian curvature $G$ can be reliably computed by the gradient and Hessian matrix of the function $\hat{f}(\cdot)$.
\begin{equation}\label{eqn:curvature_estimation_mean_and_guassian}
\begin{split}
&H = \frac{\triangledown \Hat{f} \mathbf{Hess}(\hat{f}) \triangledown \Hat{f}^T - |\triangledown \Hat{f}|^2 Trace(\mathbf{Hess}(\Hat{f})))}{2|\triangledown \Hat{f}|^3}, \\
&G = \frac{\begin{vmatrix}
	\mathbf{Hess}(\Hat{f}) ~~~ \triangledown \Hat{f}^T \\
	\triangledown \Hat{f} ~~~ \quad 0  
	\end{vmatrix}}{|\triangledown \Hat{f}|^4},
\end{split}
\end{equation}
where $\triangledown$ and $\mathbf{Hess}(\cdot)$ are the gradient and Hessian operator respectively. After that, the principal curvatures can be obtained by solving the quadratic equation of normal curvature derived constructed from the first and second fundamental forms \cite{Patrikalakis2002Shape}. That is $\kappa_{1} = H + \sqrt{H^2 - G}$ and $\kappa_{2} = H - \sqrt{H^2 - G}$.

To evaluate the reliability of curvature estimation, a comparison between the prior approach \cite{lefloch2017} based on quadratic surface fitting and our method is given in Fig.~\ref{fig:comparison_curvature_estimation}. As can be observed in the zoom-view, curvature estimation applied by~\cite{lefloch2017} is quite unstable in noisy regions (see the undefined points shown in black). In contrast, our approach based on local HRBF approximate is robust to noise. The result of curvature evaluation is stored in a map co-aligned with the vertex map $\mathbf{D}$.

\subsection{Reconstruction-indicated confidence map}
\label{subsecConfidenceMap}
For each new input RGB-D frame, a confidence map is usually constructed to indicate the level of confidence at each vertex for the camera pose estimation and the depth fusion. In the previous reconstruction systems~\cite{Keller2013, Whelan2016ElasticFusionRD, lefloch2017}, the confidence map $\Upsilon$ for each raw input is derived from the radial decreasing quality~\cite{Keller2013} according to the distortion model of the camera~\cite{sarbolandi2015kinect} -- i.e., the depth values on pixels closer to the center of the camera are more accurate. The distortion-based method can improve the reconstruction quality to some extent but it still ignores the uncertainty of the input data. Hence we evaluate the input depth map by a reconstruction-indicated method based on the observation that the implicit surface reconstruction relies on the density and reliability of the acquired points. 

The confidence is higher in regions where dense points exist to construct an implicit surface and vice versa~\cite{QualityDriven2014}. Specifically, we evaluate the magnitude of the function gradient $\triangledown \Hat{f}(\mathbf{v})$ and its consistency to the normal $\tilde{\mathbf{n}}_D$ indicated by the depth map $\mathbf{D}$. This is because a reliable local shape described by HRBF implicits will 1) be commonly defined by more kernels and 2) have its gradient pointing toward the similar direction as $\tilde{\mathbf{n}}_D$. Therefore, the reconstruction-indicated confidence can be evaluated by
\begin{equation}\label{eqn:reconstruct_confidence}
c_r =exp(-\frac{\varepsilon}{||\triangledown \Hat{f}(\mathbf{v}) \cdot \tilde{\mathbf{n}}_D||})
\end{equation}
with $\tilde{\mathbf{n}}_D$ being the unit normal obtained by applying central-difference on the bilateral filtered depth values of $\mathbf{D}$. $\varepsilon$ is a coefficient to reflect the resolution of RGB-D cameras. For all our experimental tests taken on a Microsoft Kinect v1 sensor, $\varepsilon=1000$ gives the best results. For each pixel $\mathbf{u}$, its final confidence is commonly determined by the reconstruction-indicated term $c_r$ and the distortion-based term $c_d$ as
\begin{equation}  \label{eqn:confidence_evaluation}
c =  c_r c_d,
\end{equation}
where $c_d = exp(-{\gamma^2}/{2\sigma^2})$ is the same as \cite{Keller2013}. Here $\gamma$ is the radial distance between the current pixel and the camera center normalized by the diagonal length of the frame image, and $\sigma = 0.6$ is derived empirically according to~\cite{Keller2013}.

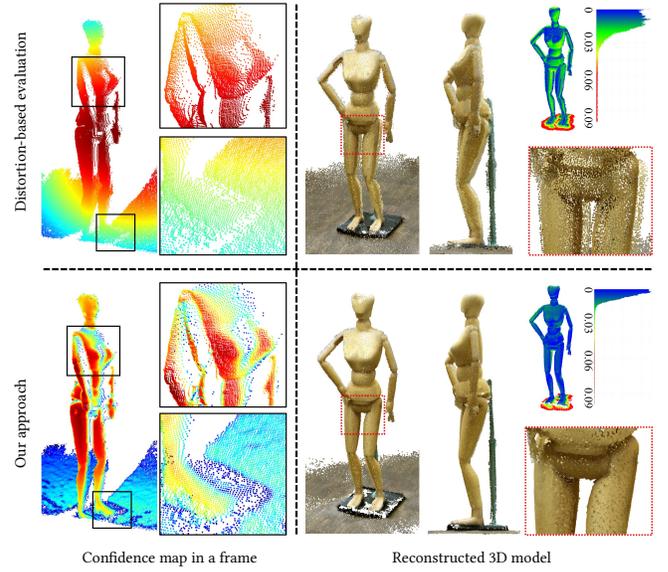
\begin{figure}[t]
	\setlength{\unitlength}{0.1\textwidth}
	\begin{picture}(5, 4.3)
	\put(0.2,0.2){\includegraphics[width=0.95\linewidth]{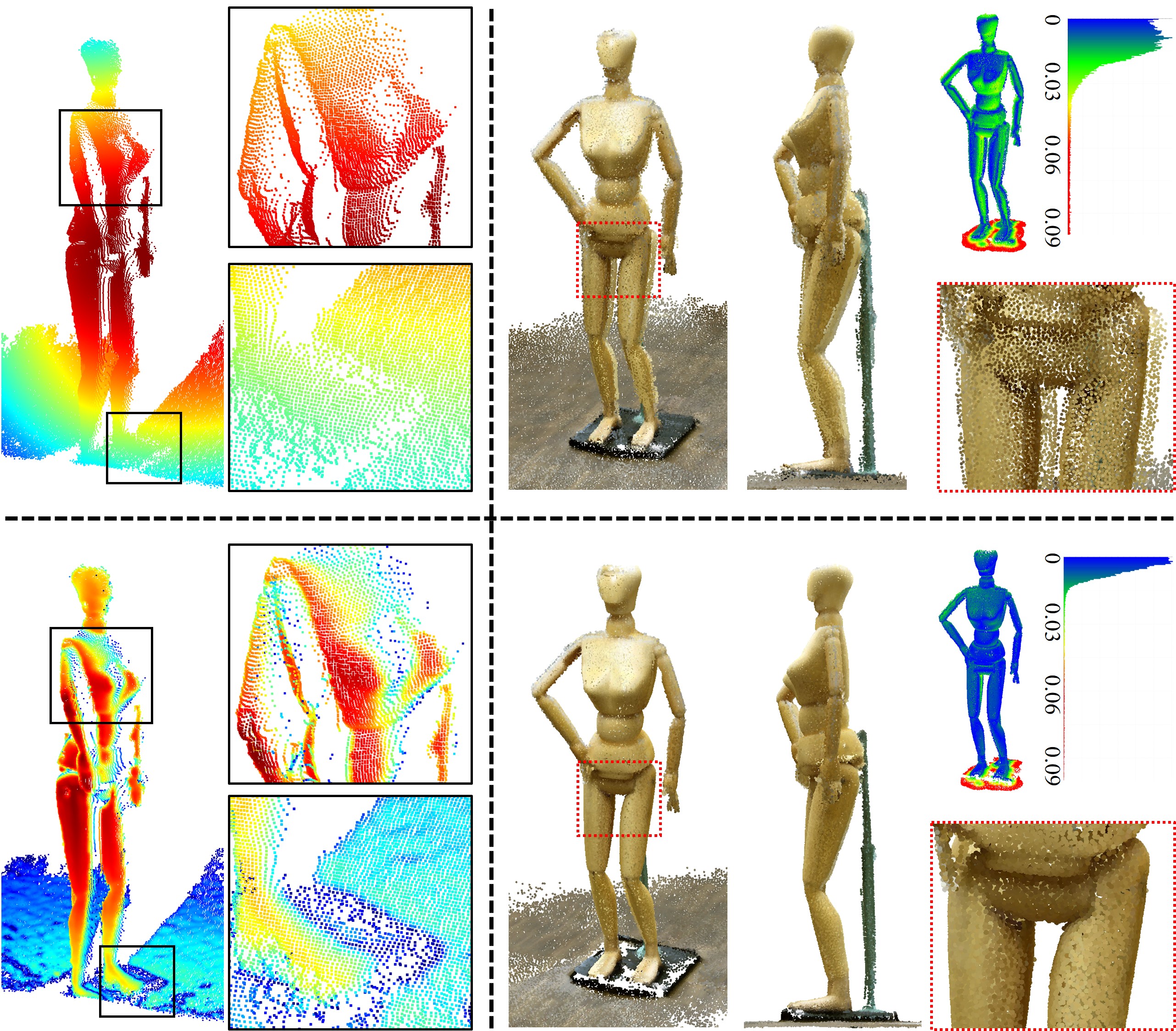}}
	\put(2.80,0){\scriptsize Reconstructed 3D model}
	\put(0.50,0){\scriptsize Confidence map in a frame}
	\put(0,2.6){\rotatebox{90}{\scriptsize Distortion-based evaluation}}
	\put(0,0.7){\rotatebox{90}{\scriptsize Our approach}}
	\end{picture}
\caption{Comparison of different methods for generating confidence maps in registration -- the camera-distortion based~\cite{Keller2013} (top left) versus our reconstruction-indicated method (bottom left), where the test is conducted on the \textit{human} model from the CoRBS benchmark~\cite{Wasenmuller2016}. The reconstructed models by using different confidence maps are shown in the right column, where zoom views highlight the quality difference in reconstruction. The reconstruction errors are measured by the distance between each point to the surface of the ground-truth model and are plotted in heat color with the corresponding histogram.
}\label{fig:compar_con_enval}
\end{figure}

We compare our method of confidence map evaluation with the camera-distortion based method~\cite{Keller2013} on the~\textit{human} model from the CoRBS benchmark~\cite{Wasenmuller2016} (Fig.~\ref{fig:compar_con_enval}). In contrast to the method of Keller et al.~\shortcite{Keller2013} that generates weights according to the optical direction of the camera, our method of confidence evaluation properly reflects the underlying uncertainty of the input frames (see the left column of Fig.~\ref{fig:compar_con_enval} for an illustration). Moreover, we further evaluate the reconstruction results by using different confidence maps as shown in the right column of Fig.~\ref{fig:compar_con_enval}. As can be found in the zoom views, misalignment occurs by using the camera-distortion based method (see the double-layers in the zoom views of Fig.~\ref{fig:compar_con_enval}'s top-right). Differently, our method can effectively reflect the unstable measurement in those regions with large depth variation by assigning smaller weight values. As a result, the registration based on our HRBF-based confidence evaluation provides better-aligned results (see the bottom-right of Fig.~\ref{fig:compar_con_enval}). We also measure the errors of reconstruction by the distance between each point to the ground-truth surface model, which are plotted in heat color with the corresponding histogram. In summary, our method leads to a more accurate 3D model with less artifact.

\section{Camera Pose Estimation}
\label{secCameraPose}
We estimate the camera pose of each newly captured RGB-D frame by registering it onto the global model, which highly depends on the underlying registration algorithm and is a key to 3D reconstruction in high accuracy. Our registration method consists of two steps:
\begin{enumerate}
\item Searching the correspondence between each point of the input frame and its corresponding point in the vertex map predicted from the global model in the previous frame;

\item Updating the registration transformation by minimizing the weighted point-to-plane geometric metric and the photometric difference between the pairs of points with correspondence determined in the first step.
\end{enumerate}
These two steps are repeatedly applied until the registration converges to obtain the relative transformation between the neighboring frames. With the help of on-the-fly HRBF surfaces proposed in our approach, curvatures and confidence maps can be reliably estimated to improve the robustness of registration. 

\subsection{Correspondence search}\label{subsecCorrespondSearch}
Given a point $\mathbf{v}_i = \mathbf{V}_i(\mathbf{u})$ from the $i$-th frame (the input RGB-D frame), it is required to find its most similar point $\mathbf{\Bar{v}}_{i-1}$ on the global model in the $(i-1)$-th frame's vertex map predicted by on-the-fly HRBF implicits. Assuming the motion between two consecutive frames is very small, the projective data association algorithm~\cite{Blais1995} can be applied to speed up the search of correspondence (ref.~\cite{Keller2013, Newcombe2011}). Specifically, the estimated transformation $\mathbf{T}_i$ to the global model, which is initialized as $\mathbf{T}_{i-1}$ and will be updated during the iteration of registration, is 
used to transfer 3D points of the $i$-th frame into the previous frame by $\mathbf{T}_{i-1}^{-1}\mathbf{T}_i$. After that, we use a small window with a fixed size of $5 \times 5$ to search compatible points in the predicted vertex map of the global model. 

We measure the dissimilarity of a point pair using the following metric similar to \cite{lefloch2017}
\begin{equation}  \label{eqn:curvature_enhancement_icp_disimilarity}
\gamma_d  = \mu_{d} I_{d} + \mu_{a} I_{a} + \mu_{c} I_{c},
\end{equation}
which is determined by the distance variation term $I_{d}$, the angle variation term $I_{a}$ and the curvature variation term $I_{c}$ together with equal weight (i.e., $\mu_d=\mu_a=\mu_c=\frac{1}{3}$ works well in all our experiments). 
\begin{equation}  \label{eqn:curvature_enhancement_icp_disimilarity_terms}
\begin{split}
&I_{d} = {\|\mathbf{v}_i -\mathbf{\bar{v}}_{m} \|}/{R_{\max}}, \\
&I_{a} = 1 - \mathbf{n}_i \cdot \mathbf{\bar{n}}_m / (\| \mathbf{n}_i\| \| \mathbf{\bar{n}}_m \|), \\
&I_{c} = 1 - exp\left(- \frac{|\mathbf{\kappa}_{1, i} - \mathbf{\kappa}_{1, m}| + |\mathbf{\kappa}_{2, i} - \mathbf{\kappa}_{2, m}|}{\max\{|\kappa_{1,m}|,|\kappa_{2,m}|\}} \right),
\end{split}
\end{equation}
where $\mathbf{\bar{v}}_{m}$, $\mathbf{\bar{n}}_m$, $\mathbf{\kappa}_{1, m}$ and $\mathbf{\kappa}_{2, m}$ are from the candidate points obtained from the global model. $R_{\max}$ is the distance between $\mathbf{v}_i$ and the farthest point that can be found in the search window. 

A point pair with the smallest values of $\gamma_d$ is considered as the valid corresponding points. Moreover, we apply a pruning strategy similar to \cite{Newcombe2011} to discard outliers in the correspondence pairs. 
A reliable correspondence search depends on a robust normal and curvature estimation, which has been improved by using our on-the-fly HRBF surface evaluation. The pairs of compatible points are stored in a set $\Psi=\{(\mathbf{u}, \mathbf{\bar{u}})\}$ for computing the updated transformation $\mathbf{T}_i$.

\subsection{Transformation update}\label{subsecRegistration}
The transformation is updated by minimizing an objective function considering both geometric and photometric information.

\subsubsection{Geometric term}~~A curvature-based weight scheme \shortcite{lefloch2017} is employed here to enhance the point-to-plane metric~\cite{Newcombe2011} for aligning an input RGB-D frame to the global model. The objective function to be minimized is defined as 
\begin{equation} \label{eqn:optimization_point2plane_metric_weighted}
E_{geom}(\mathbf{T}_{i}) = \sum_{(\mathbf{u}, \mathbf{\bar{u}}) \in \Psi} w(\mathbf{\bar{u}}) \left( ( \mathbf{T}_{i} \mathbf{v}_i - \mathbf{\bar{v}}_{i-1} ) \cdot \mathbf{\Bar{n}}_{ i-1} \right)^2,
\end{equation}
where the curvature-based scheme \cite{lefloch2017} is employed to determine the weight $w(\mathbf{\bar{u}})$ by incorporating the confidence coefficient, the depth-value and most importantly the principal curvatures at the point $\mathbf{\Bar{v}}_{i-1}(\mathbf{\bar{u}})$. 
\begin{figure}[t]
	\centering\vspace{5pt}
	\includegraphics[width=0.95\linewidth]{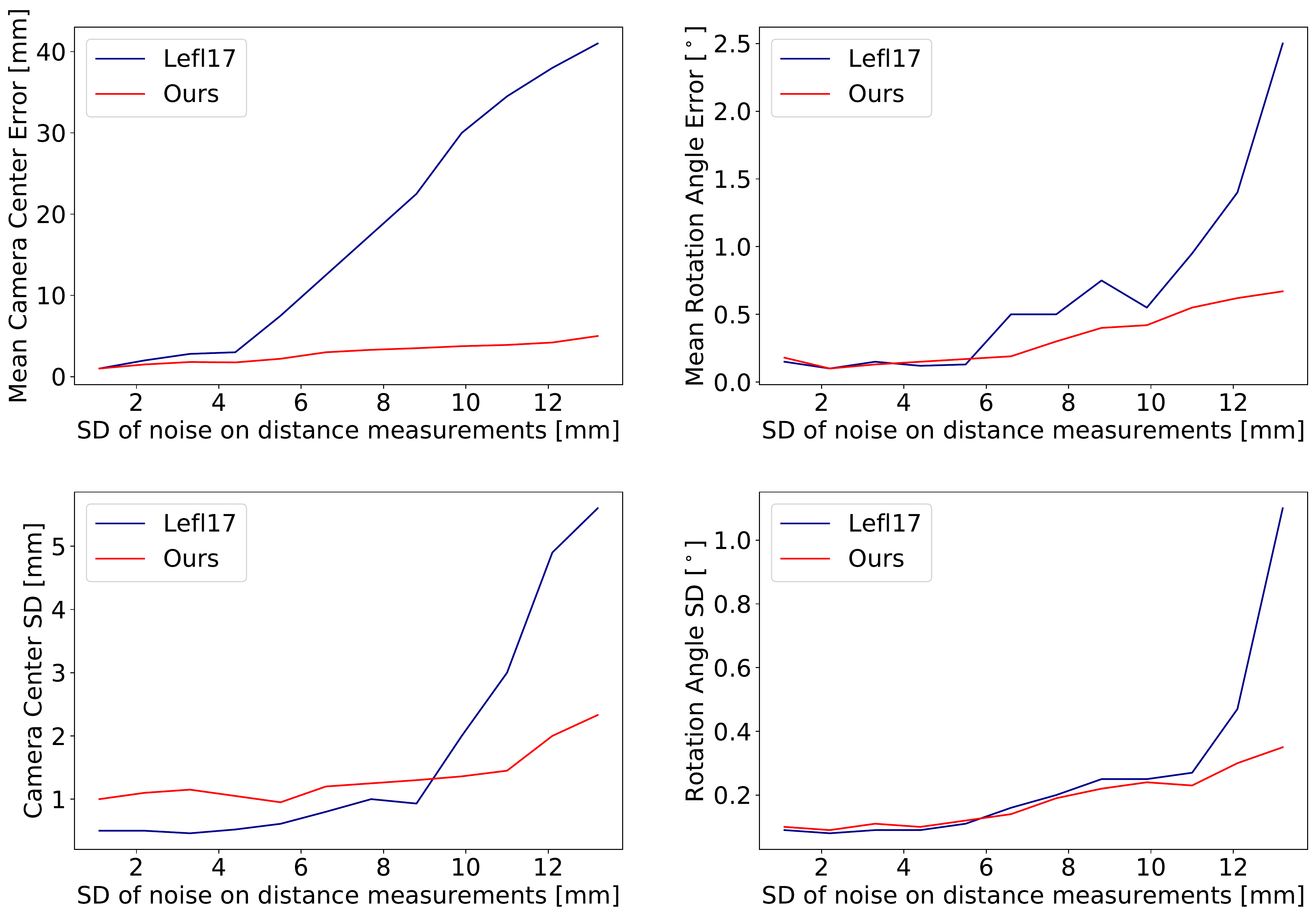}
	\caption{Comparison of tracking robustness on the \textit{Lego-PAMI-TT Noise  Benchmark} \cite{lefloch2017}. The mean error (top) and standard deviation for the estimated camera center (bottom-left) and rotation (bottom-right) are evaluated with different levels of (Gaussian) noise. The noisy levels successively increases with standard deviation by integer factors (i.e., 1-13). Note that the camera-distortion based confidence maps~\cite{Keller2013} are employed in \cite{lefloch2017}.}\label{noise_quantitative_evaluation}
\end{figure}
Figure~\ref{noise_quantitative_evaluation} demonstrates the performance improvement when using on-the-fly HRBF to evaluate curvatures (Section \ref{subsecCurvature}) and confidence map (Section \ref{subsecConfidenceMap}) as proposed in this paper. In this experiment, the routine and the weighting scheme of registration are the same as \cite{lefloch2017}. We evaluate the camera
tracking accuracy by computing Mean Camera Center Error and Standard Deviation (SD) between the estimated poses with the corresponding reference poses, as described in~\cite{lefloch2017}. This comparison indicates that our method significantly improves the accuracy of registration (therefore camera pose estimation) when high-level noise is presented.

\subsubsection{Photometric term}~~Following the approach of ElasticFusion~\cite{Whelan2016ElasticFusionRD}, color information provided by an RGB-D camera is used to further enhance registration. 
This complementary information is encoded in a photometric term as
\begin{equation}
E_{color}(\mathbf{T}_{i}) = \sum_{(\mathbf{u}, \mathbf{\bar{u}}) \in \Psi}(\bar{\mathbf{C}}_{i-1}(\pi(\mathbf{T}^{-1}_{i-1}\mathbf{T}_{i}\mathbf{v}_i))- \mathbf{C}_{i}(\mathbf{u}))^2,
\end{equation}
where $\mathbf{\bar{C}}_{i-1}$ and $\mathbf{C}_i$ denote the RGB color value in the predicted map of the previous frame and the color in the current input frame. $\pi$ is the projection function between 3D objects and the corresponding image frame. 

The final objective function to be minimized is
\begin{equation}
\label{eqn:align}
E(\mathbf{T}_{i}) =w_{geom}E_{geom}(\mathbf{T}_{i}) + E_{color}(\mathbf{T}_{i}),
\end{equation}
where $w_{geom}$ is the weight of the geometric term. $w_{geom}=10$ is suggested in \cite{Whelan2016ElasticFusionRD} and works well in all our tests. We employ the Gauss-Newton nonlinear least-squares method~\cite{bjorck1996numerical} to minimize this energy function, which leads to a reliable alignment between the current input frame and the global model usually after around $20$ steps of iteration.

\section{Depth Map Fusion}
\label{secDepthFusion}
Given a valid camera pose (Section \ref{secCameraPose}), the depth map fusion step integrates the input points and their attributes into a global model as an enriched surfel representation (Section \ref{secOverview}). 

Let $\mathbf{T}_{i} \in \mathbb{SE}_3$ denote the pose of $i$-th input frame, we transform both points and their normals into the $(i-1)$-th frame to conduct the data fusion. Points of the global model are also projected into the $(i-1)$-th frame with their vertex ID stored in the texture map. After that, for each transformed point of the $i$-th frame, we follow the scheme of \cite{Cao2018,Keller2013} to search its \textit{valid} neighbors in a $5 \times 5$ window by using the same position / normal compatibility condition. When there are multiple \textit{valid} neighbors, the closest one is chosen to conduct fusion by a confidence-weighted averaging. That is,
\begin{equation} \label{eqn:fusion_vnc}
\begin{gathered}[t]
\bar{\mathbf{v}} \gets \frac{\bar{c}\bar{\mathbf{v}} + c \mathbf{v}_{g,i}}{\bar{c} + c},
\bar{\mathbf{n}} \gets \frac{\bar{c}\bar{\mathbf{n}} + c \mathbf{n}_{g,i}}{\bar{c} + c},  \\
\bar{\kappa}_{1} \gets \frac{\bar{c}\bar{\kappa}_{1} + c\kappa_{1, i}}{\bar{c} + c}, \bar{\kappa}_{2} \gets \frac{\bar{c}\bar{\kappa}_{2} + c\kappa_{2, i}}{\bar{c} + c}, \\
\bar{r} \gets \bar{c}\Bar{r} + c r, \Bar{c} \gets \bar{c} + c, \bar{t} \gets t_i,  
\end{gathered}
\end{equation}
with $\mathbf{v}_{g,i}$ and $\mathbf{n}_{g,i}$ being the position and normal of an input point in the global model's coordinate. $\bar{c}$, $\bar{\kappa}_{1}$ and $\bar{\kappa}_{2}$ are the stored confidence and curvature values of a point on the global model, and $c$, $\kappa_{1, i}$ and $\kappa_{2, i}$ are the values on an input point evaluated by using the on-the-fly HRBF implicits (Section \ref{subsecCurvature} and \ref{subsecConfidenceMap}). 

Points of global model with confidence above a threshold $\sigma_{conf}$  are considered as \textit{stable} points (e.g., $\sigma_{conf}=5.0$ is employed by following \cite{Cao2018,Keller2013}), and only stable points are used for HRBF surface prediction (Section \ref{subsecSurfaceEvaluation}). If no corresponding model point is identified, we add the current vertex with its attributes to the global model as an unstable point. Besides, we remove points with confidence values below this threshold for a period of time (i.e., 200 frames) by considering them as noises or outliers.

\section{Results and Discussion}
\label{secResult}
\begin{figure}[t]
\includegraphics[width=\linewidth]{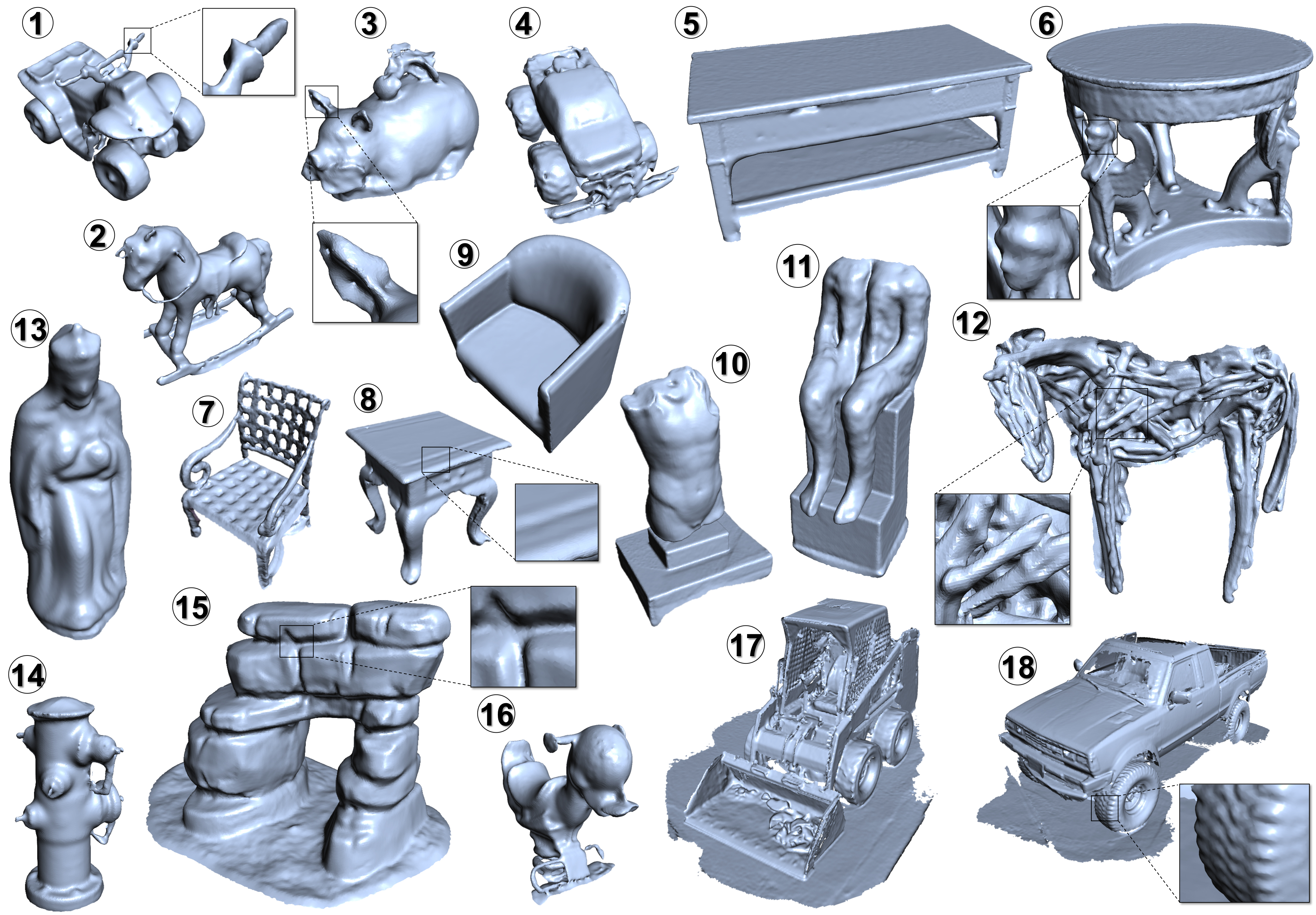}
\caption{Reconstructed 3D individual models with different geometric properties and scalability from the~\textit{Object Scans}~\cite{Choi2016} dataset.
Here the models are rendered by polygonal meshes extracted from the iso-surfaces of HRBF implicits.}
\label{fig:reconstruction_3Dscene}
\end{figure}

We have implemented our algorithm\footnote{The source code is available at: \url{https://github.com/YabinXuTUD/HRBFFusion3D}.} in the framework of ElasticFusion~\cite{Whelan2016ElasticFusionRD} by C++, CUDA, and OpenGL Shading Language. Moreover, we have incorporated \textit{ORB-SLAM2}~\cite{Artal2017} into our system for the implementation of submap-based hierarchical optimization for large-scale scanning. Our system has been evaluated on both synthetic datasets and raw sequences captured by various depth cameras, including structured light cameras (e.g., \textit{Asus XTion PRO LIVE}, \textit{PrimeSense Carmine} and \textit{Microsoft Kinect v1}) as well as Time-of-flight cameras (e.g., \textit{Microsoft Kinect v2}). We carried out all our experiments on a desktop PC equipped with an Intel Core i7-9700K CPU @3.60GHz with 16GB RAM and a GeForce RTX 2070 GPU with 8GB memory. 
In this section, we first briefly describe the datasets. Then we present our visual results, followed by the evaluation of our method on different datasets. The output of our system can be either a point cloud or a triangular mesh extracted from the iso-surface maintained by the closed-form HRBF representation using the dual-contouring method~\cite{liu2016}. Figure~\ref{fig:reconstruction_3Dscene} shows some small and middle-sized objects rendered in meshes. While in all other figures, we directly render point clouds for the sake of efficiency. All reconstructed 3D models are visualized by Easy3D~\cite{easy3d2021}, which is an open-source library for 3D modeling, geometry processing, and rendering.

\subsection{Test datasets}\label{subsecTestDataset}
We tested our method on the following datasets.

\subsubsection{Object Scans~\cite{Choi2016}} 
This dataset provides more than 10,000 individual 3D object scans that contain a diversity of objects with different geometric properties and scalability. The scans are captured by unprofessional operators with a PrimeSense Carmine RGB-D camera.


\subsubsection{ICL-NUIM benchmark~\cite{handa:etal:ICRA2014}}
This is a synthetic benchmark dataset with geometric and camera pose ground truth. We selected four scenes of~\textit{living rooms} (including synthetic noise) commonly used in the previous work to evaluate the tracking accuracy and reconstruction quality of our results.

\subsubsection{TUM benchmark~\cite{sturm2012benchmark}}
It is a dataset captured by a \textit{Microsoft Kinect v1} with motion-captured camera poses as ground truth, which is widely used to evaluate the tracking accuracy of a reconstruction method. We select four frequently used sequences (i.e., \textit{fr1/desk}, \textit{fr2/xyz}, \textit{fr3/office}, \textit{fr3/nst}) for the evaluation.

\subsubsection{CoRBS benchmark~\cite{Wasenmuller2016}}
This is a benchmark dataset of \textit{Microsoft Kinect v2} providing both the motion-captured camera poses and the 3D models acquired by a high-precision commercial scanner as ground-truth. We select the~\textit{human} model (Fig.~\ref{fig:compar_con_enval}) and the~\textit{racing car} model (in supplementary video) to demonstrate the performance of our approach.

\subsubsection{CuFusion dataset~\cite{zhang2017cufusion}}
This dataset contains both synthetic and real-world sequences for object scanning. Both ground truth trajectories and 3D models are provided on the synthetic examples. We select the synthetic sequence~\textit{Armadillo} (that does not have color information) for the evaluation.

\subsubsection{ScanNet dataset~\cite{ScanNetDai2017}}
This dataset is an RGB-D video dataset captured by structure sensors, which consists of 2.5 million views in more than 1,500 scanned sequences. We randomly selected 200 sequences to test the performance of our approach.

\subsubsection{Our dataset}
We scanned a few objects and large indoor scenes using a \textit{Microsoft Kinect v1} and are shown in Figs.~\ref{fig:Comparison_Redwood_BundleFusion_object_reconstruction}, \ref{fig:Redwood_ElasticFusion_BundleFusion_indoor_scene}, \ref{fig:comparison_reconstruction_accuracy}, \ref{fig:evaluationOutdoorScene}, and~\ref{fig:evaluationSingleComponent}. This is mainly used to evaluate the detail recovery and the scalability of our method. 
For the evaluation of reconstruction accuracy, we obtain the ground truth models shown in Fig.~\ref{fig:comparison_reconstruction_accuracy} by a commercial hand-held structure light scanner, Artec Eva, with the precision of $0.1mm$.

\subsection{Visual results}
\subsubsection{Individual objects}
We first tested our method on a variety of objects from the \textit{Object Scans} dataset~\cite{Choi2016}. Figure \ref{fig:reconstruction_3Dscene} shows the reconstruction results of 18 objects of different sizes and characteristics. Among these objects, (1), (2), (3), (4) are small toys, where the average size is about $0.56m \times 0.30m \times 0.36m$, small geometric features are presented (i.e., the handlebar in (1) with a radius of $0.01m$, the ear in (3) with a thickness of $0.02m$~(as shown in zoom-views of Fig.~\ref{fig:reconstruction_3Dscene}). It is intractable for the methods based on a volumetric representation to reconstruct such geometric details while still adapting to the scale of its background. On the contrary, our HRBF-based on-the-fly surface representation has addressed such limitation since the reconstruction quality only depends on the local kernels and the corresponding support radius (Section \ref{subsecSurfaceEvaluation}). 

Apart from the small-sized toys, we also tested our system on middle-sized objects, including indoor furniture (5) (6) (7) (8) (9), sculpture (10) (11) (12) (13), and outdoor equipment (14) (15) (16). The average size is around $0.89m \times 0.60m \times 1.03m$. The chair (9), the sculpture (10) (11), and the outdoor equipment (15) mainly demonstrate curved surfaces while the tables (5) (6) (8) contain large planar regions. Besides, the chair (7) and the horse models (12) have dense tube-like structures, which poses challenges for RGB-D reconstruction systems. Thanks to the high adaptivity provided by the HRBF on-the-fly surface representation, such fine geometric features (i.e., the decoration on the legs of table (6), the small crease on the desktop in (8), and the concave part in (15) (see the zoom-views in Fig.~\ref{fig:reconstruction_3Dscene})) are faithfully recovered by our system. 

At last, we tested our system with relatively large vehicles (17) and (18), the sizes of which are $4.94m \times 1.97m \times 1.94m$ and $3.33m \times 1.68m \times 1.94m$ respectively. Our system can reconstruct not only global consistent models but also fine geometric details. This can be observed from the crease of the tires on both objects (as shown in the zoom-view in Fig.~\ref{fig:reconstruction_3Dscene}). 

\subsubsection{Large scenes}
Figure \ref{figTeaser} presents two large-scale indoor scenes reconstructed by our system. The left shows the reconstructed results of a study room in a university library, while the right shows a study platform in a grand hall of an academic building.
Please note that the length of both scenes is above $21m$. Due to the complexity of the scene layout, the camera trajectories are extremely complicated, posing challenges to both camera tracking and reconstruction. The detailed camera trajectories can be found in our supplementary video. Our system managed to capture and reconstruct both scenes with high fidelity.

\subsection{Evaluation}
In addition to the above visual results, we also conducted a comprehensive analysis of our method in terms of tracking robustness, detail recovery, scalability, reconstruction accuracy, ablation study, parameter discussion, memory consumption, and processing times. Details are given below.

\subsubsection{Tracking robustness}
\label{subsecResTrackingRobustness}
The camera tracking of RGB-D reconstruction systems generally tends to drift due to the noise and sparsity in the input frames, which also accumulates noise in the global model. We evaluated the performance of our HRBF-based surface evaluation in tracking robustness below. 

\begin{figure}[t]
\centering
\includegraphics[width=\linewidth]{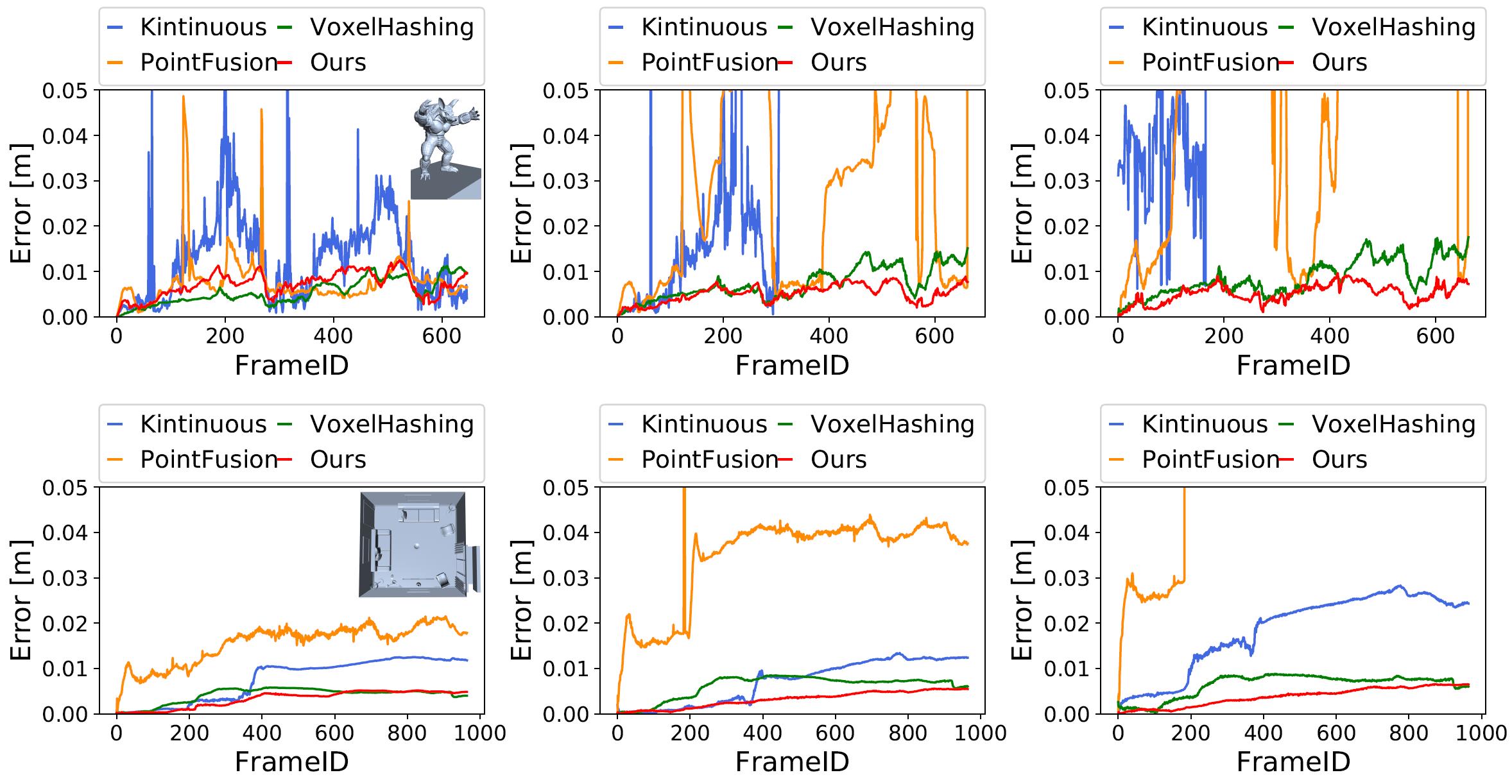}
\vspace{-15pt}
\caption{Comparison of the influence of different representations on tracking robustness with \textit{Kintinuous}~\cite{whelan2012kintinuous}, \textit{PointFusion}~\cite{Keller2013} and \textit{VoxelHashing}~\cite{NieBner2013} on two noisy scanning sequences -- (top row) the \textit{Armadillo} model of the \textit{CuFusion Dataset}~\cite{zhang2017cufusion} and (bottom row) the scene sequence \textit{lr kt1} of the \textit{ICL-NUIM dataset}~\cite{handa:etal:ICRA2014}. 
From left to right, noises are added into the depth maps in different levels of normal distribution: $\sigma=3.0$, $\sigma=6.0$ and $\sigma=12.0$. 
Note that we clip tracking error larger than 0.05m and consider it as tracking lost. The insets show the ground truth-geometry of test data.}
\label{fig:quantitative_evaluation_icl_nuim}
\end{figure}

Three state-of-the-art reconstruction systems are selected to compare the influence of different representations on tracking robustness, including \textit{Kintinuous}~\cite{whelan2012kintinuous}, \textit{PointFusion}~\cite{Keller2013}, and \textit{VoxelHashing}~\cite{NieBner2013}. \textit{Kintinuous} is an extended version of the original \textit{KinectFusion}~\cite{Newcombe2011} by exploiting a dynamic volume. \textit{VoxelHashing} utilizes a hashing structure to maintain a sparse representation with voxel grids. \textit{PointFusion} uses a surfel representation for camera tracking. The experiment is conducted on two synthetic sequences with ground truth camera poses: the \textit{Armadillo} of the \textit{CuFusion Dataset}~\cite{zhang2017cufusion} and the \textit{lr kt1} of the \textit{ICL-NUIM dataset}~\cite{handa:etal:ICRA2014}. Noises are added in different levels of normal distribution (i.e., $\sigma=3.0$, $\sigma=6.0$ and $\sigma=12.0$) to test the robustness of different systems.  We evaluated the camera pose error for all frames and the results are shown in Fig.~\ref{fig:quantitative_evaluation_icl_nuim}. It can be found that the point-based representation is more sensitive to noise while our HRBF-based method demonstrates consistently low errors in camera tracking. 

\begin{figure*}[htbp]
	\setlength{\unitlength}{0.1\textwidth}
	\begin{picture}(10,7.1)
	\put(0.35,0.04){\includegraphics[width=0.95\textwidth]{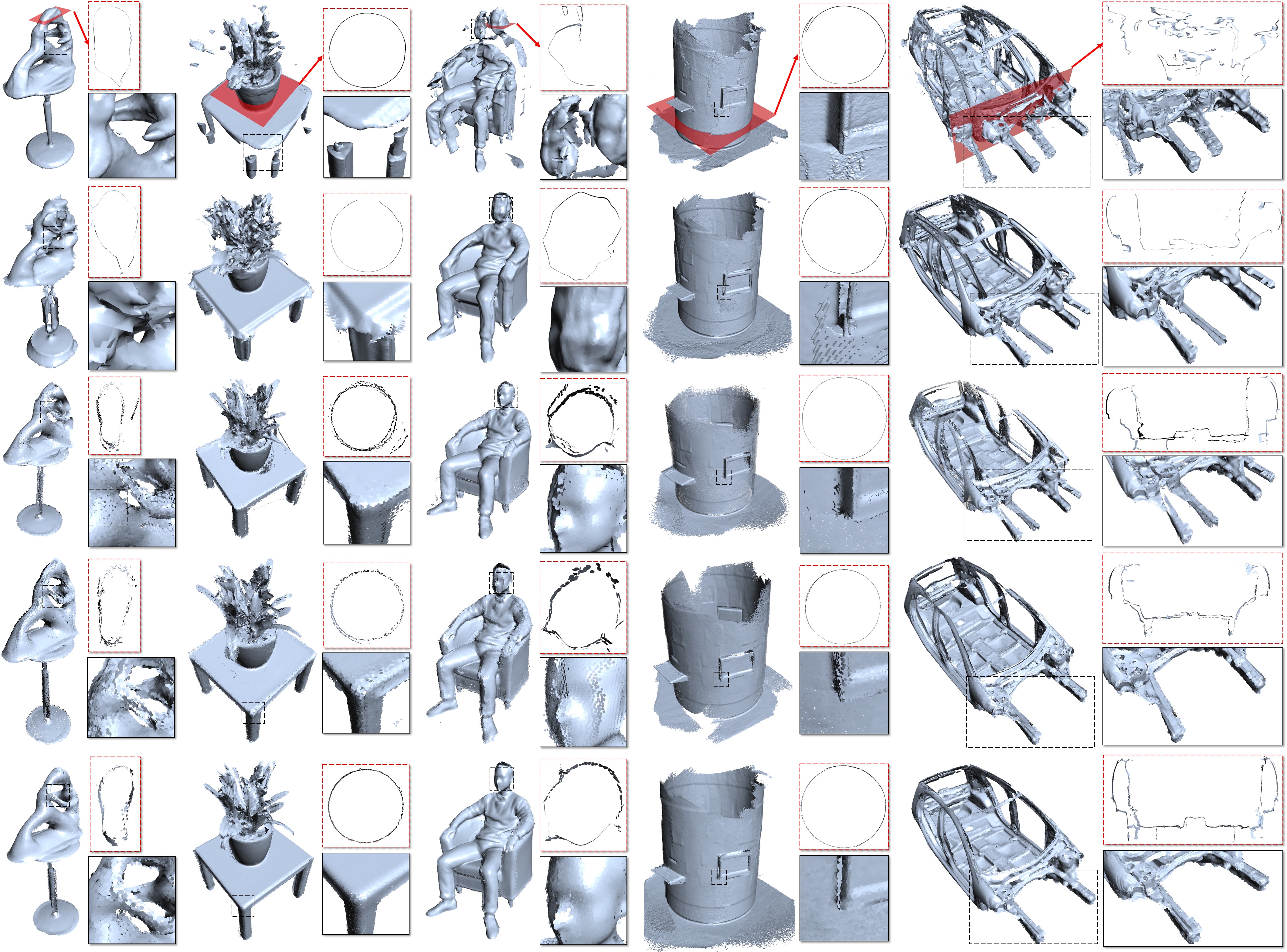}}
	\put(0.18, 0.6){\scriptsize \rotatebox{90}{Ours}}
	\put(0.18, 1.7){\scriptsize \rotatebox{90}{UncertaintyAware}}
	\put(0.18, 3.3){\scriptsize \rotatebox{90}{ElasticFusion}}
	\put(0.18, 4.7){\scriptsize \rotatebox{90}{BundleFusion}}
	\put(0.18, 6.1){\scriptsize \rotatebox{90}{Redwood}}
	\end{picture}
	\caption{Comparison of reconstruction results generated by \textit{Redwood}~\cite{Choi2016}, \textit{BundleFusion}~\cite{Dai2017}, \textit{ElasticFusion}~\cite{Whelan2016ElasticFusionRD},  \textit{UncertaintyAware}~\cite{Cao2018}, and our method on five objects with different geometric shapes and details. Models from left to right are: \textit{Fertility}, \textit{Plant}, \textit{Human}, \textit{Pillar} and \textit{Car Frame}. Note that the \textit{Redwood} and the \textit{BundleFusion} methods generates mesh surfaces from volume representation as results (displayed in the first two rows) while the results of other three methods as point clouds are rendered by surfel splatting.
	}\label{fig:Comparison_Redwood_BundleFusion_object_reconstruction}
\end{figure*}


\begin{table}[t]
	\caption{ATE RMSE on the TUM benchmark (unit: m)}
	\label{table:TUM_ate_rmse}\vspace{-8pt}\centering\small
			\begin{tabular}{c|c|c|c|c}
				\hline
   & \textit{fr1/desk} & \textit{fr2/xyz} & \textit{fr3/office} & \textit{fr3/nst} \\
				\hline
				\hline
				\textit{DVO} SLAM & 0.021 & 0.018 & 0.035 & 0.018  \\
				\hline
				\textit{RGBD SLAM} & 0.023 & 0.008 & 0.032 & 0.017  \\
				\hline
				\textit{MRSMap} & 0.043 & 0.020 & 0.042 & 2.018  \\
				\hline
				\textit{Kintinuous} & 0.037 & 0.029 & 0.030 & 0.031  \\
				\hline
				\textit{ElasticFusion} & 0.020 & 0.011 & 0.017 & 0.016  \\
				\hline
		\textit{BundleFusion} & 0.016 & 0.011 & 0.022 & \textbf{0.012}  \\
				\hline
		\textit{Redwood} & 0.027 & 0.091 & 0.030 & 1.929  \\
				\hline
		\textit{UncertaintyAware} & {0.015} &{0.006}  & {0.009} &  {0.014}  \\
				\hline
		\textit{Ours} & \textbf{0.014} & \textbf{0.005} & \textbf{0.007} & 0.016   \\
				\hline
				\hline
	\multicolumn{5}{c}{{Comparison of only applying Local BA}}			\\
				\hline
		\textit{{UncertaintyAware}} & {0.015} & {0.006}  & {0.037} & {0.014}  \\
				\hline
		\textit{{Ours}} & {0.014} & {0.005} & {0.015} & {0.016}   \\
				\hline
				\hline
	\multicolumn{5}{c}{{Comparison of only applying Global BA}}			\\
				\hline
		\textit{{UncertaintyAware}} & {0.033} & {0.009}  & {0.025} & {0.093}  \\
				\hline
		\textit{{Ours}} & {0.018} & {0.007} & {0.014} & {0.030}   \\
				\hline
			\end{tabular}
			\vspace{-10pt}
\end{table}

We further evaluated our system in terms of accuracy in trajectory estimation on the \textit{TUM benchmark}~\cite{sturm2012benchmark} (\textit{Microsoft Kinect v1}) where ground truth trajectories are provided by a highly accurate calibrated motion-capture system. We chose a set of widely used sequences (i.e, \textit{fr1/desk}, \textit{fr2/xyz}, \textit{fr3/office}, \textit{fr3/nst}) and compared our methods with state-of-the-art online reconstruction systems, including \textit{DVO-SLAM}~\cite{kerl2013dense}, \textit{RGBD SLAM}~\cite{endres2012evaluation}, \textit{MRSMap}~\cite{stuckler2014multi}, \textit{Kintinuous}~\cite{whelan2012kintinuous}, 
\textit{ElasticFusion}~\cite{Whelan2016ElasticFusionRD}, 
\textit{BundleFusion}~\cite{Dai2017}, and \textit{UncertaintyAware}~\cite{Cao2018}. To make a complete comparison, the offline reconstruction system, \textit{Redwood}~\cite{choi2015robust}, is also included. We recorded the absolute trajectory error (ATE) of root-mean-square error (RMSE) for camera tracking accuracy. 
The results are summarized in Table~\ref{table:TUM_ate_rmse}. We can see that our method consistently outperformed (or demonstrated comparable) results to the most promising methods in the comparison. To analyze camera tracking drift, we separately evaluated our method with and without global optimization (i.e., similar to \textit{UncertaintyAware}~\cite{Cao2018} that both local and global bundle adjustment (BA) are applied in global optimization). Most existing systems have applied different global optimization techniques to alleviate the accumulated errors in camera pose estimation.
\begin{itemize}
    \item \textit{DVO SLAM}, \textit{RGBD SLAM}, and \textit{MRSMap} first apply a pose graph optimization to achieve a global consistent trajectory and then the global model is constructed by integrating all depth maps in a volumetric representation (i.e., \textit{DVO SLAM} and \textit{RGBD SLAM}) or merging key surfel views (i.e., \textit{MRSMap}).
    
    \item \textit{Kintinuous} and \textit{ElasticFusion} achieve a globally consistent model in a map-centric manner by deforming the global model according to global or local constraints.
    
    \item \textit{Redwood}, \textit{BundleFusion}, and \textit{UncertaintyAware} divide the global model into submaps and obtain a globally consistent model by optimizing between submaps. 
\end{itemize}
Our system outperforms most of these systems. 
There is one exception that \textit{BundleFusion} achieved the best result on \textit{fr3/nst}. The main reason lies in its combined sparse visual features, dense photometric and geometric objective, which enables it to obtain a more tight alignment on textured scenes. Global optimization techniques such as local or global BA can help significantly reduce the tracking errors in practice. It is interesting to compare the errors after removing either local or global BA (see the last two parts of Table \ref{table:TUM_ate_rmse}). We can find that our results are more accurate than those of \textit{UncertaintyAware} in most cases. 



\subsubsection{Detail recovery} 
With the help of on-the-fly HRBF surface representation, our method is able to recover finer geometric details. To demonstrate this capability, we compared our method with the state-of-the-art reconstruction systems including \textit{Redwood}~\cite{choi2015robust}, \textit{ElasticFusion}~\cite{Whelan2016ElasticFusionRD}, \textit{BundleFusion}~\cite{Dai2017}, and \textit{UncertaintyAware}~\cite{Cao2018} on a variety of 3D objects (see Fig.\ref{fig:Comparison_Redwood_BundleFusion_object_reconstruction}). Since our scanning aims at achieving a complete model, a global loop is required to exist for every model. 

The \textit{Redwood} system~\cite{choi2015robust} cannot generate good results due to the registration error. It completely failed on the human example (see the human face and the right leg in the third column). The same issue of camera tracking drift also occurs in the \textit{ElasticFusion} and the \textit{BundleFusion} systems. In short, all these three systems are unable to produce a global consistent 3D model. The \textit{UncertaintyAware} approach can obtain global consistent models by successfully detecting the close loop in all examples. However, artifacts are still generated by the~\textit{UncertaintyAware} approach due to the accumulated error -- see the human face in the third column. As has been expected, our HRBF-based method is more robust in recovering geometric details.

\begin{figure}[htbp]
 \setlength{\unitlength}{0.1\textwidth}
 \begin{center}
 \begin{picture}(5, 4.35)
 	\put(0,0.15){\includegraphics[width=\linewidth]{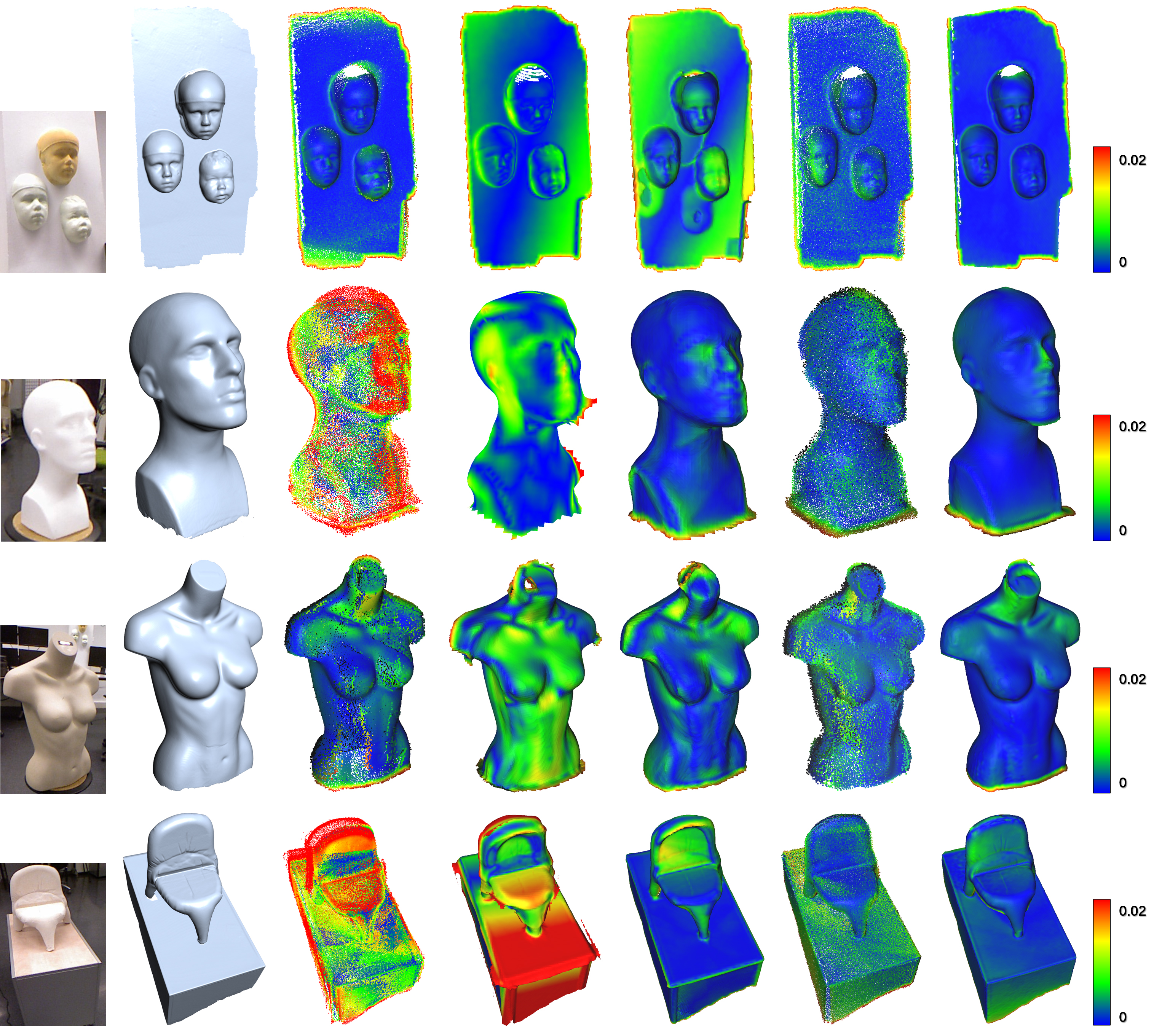}}	
 	\put(0.06,0.05){\scriptsize Image}
 	\put(0.47,0.05){\scriptsize Ground truth}
 	\put(1.17,0.05){\scriptsize ElasticFusion}
 	\put(1.87,0.05){\scriptsize BundleFusion}
 	\put(2.63,0.05){\scriptsize Redwood}
 	\put(3.13,0.05){\scriptsize UncertaintyAware}
 	\put(4.10,0.05){\scriptsize Ours}
 	\end{picture}
 \end{center}
 \caption{Comparison of the reconstruction accuracy with \textit{ElasticFusion}~\cite{Whelan2016ElasticFusionRD} (third column), \textit{BundleFusion}~\cite{Dai2017} (fourth column), \textit{Redwood}~\cite{choi2015robust} (fifth column), and \textit{UncertaintyAware} (sixth column). The ground-truth models (second column) were obtained by a high-precision structure light 3D scanner. The color map presents the distance error on the reconstructed models. Models from top to bottom are: \textit{Faces}, \textit{Head}, \textit{Upper Body}, \textit{Small Chair}.}
 \label{fig:comparison_reconstruction_accuracy}
\end{figure}

\begin{figure}
\centering
\includegraphics[width=\linewidth]{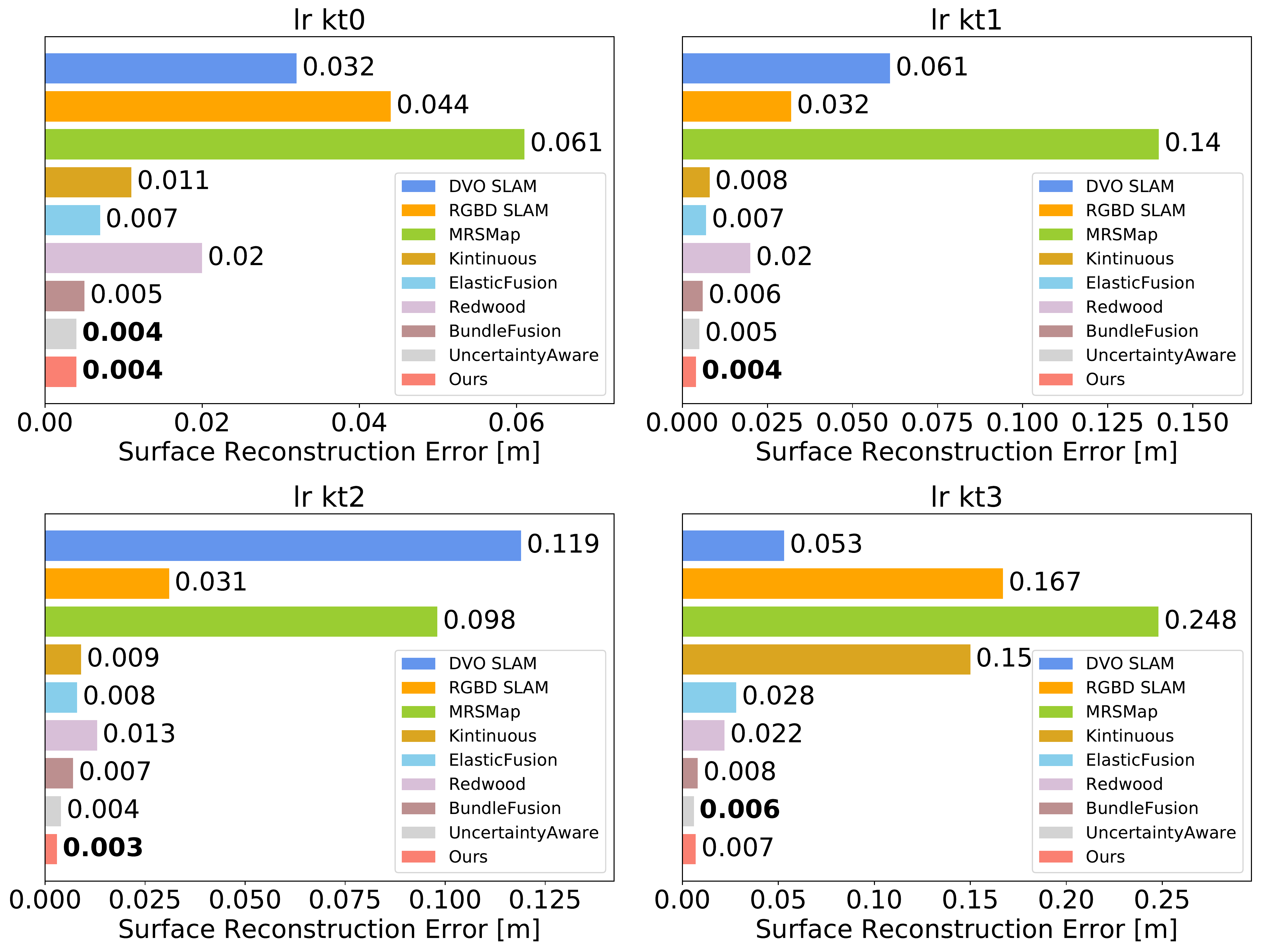}
\caption{Surface reconstruction error in terms of average point to surface distances on the \textit{ICL-NUIM benchmark} (unit: meter). The best performance is highlighted in bold fonts.}\label{fig:ICL-NUIM_surface_reconstruction_error}
\end{figure}

\begin{figure}[htbp]
    \setlength{\unitlength}{0.1\textwidth}
    \begin{picture}(10.0, 2.9)
    \put(0.3,-0.1){\includegraphics[width=0.95\linewidth]{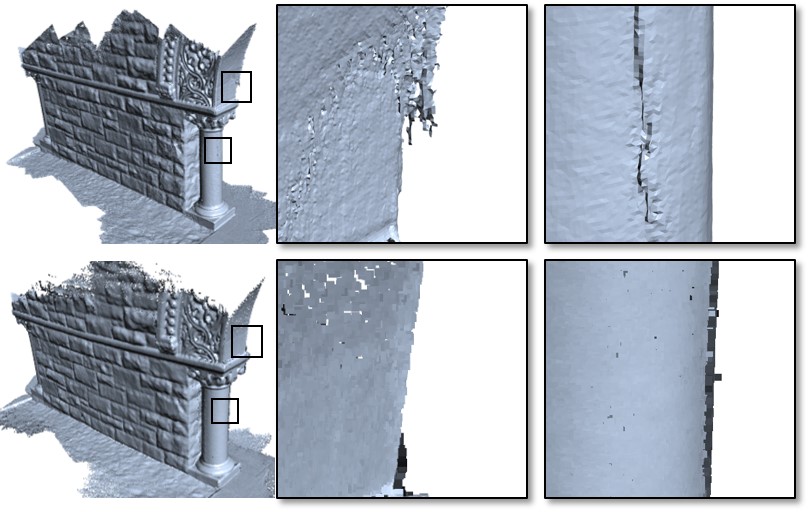}}
    \put(0.1,1.6){\scriptsize \rotatebox{90}{Zhou et al.~\shortcite{Zhou:2013}}}
    \put(0.1,0.5){\scriptsize \rotatebox{90}{Ours}}
    \end{picture}
    \caption{Comparison of the reconstruction quality between the offline optimization based method of Zhou et al.~\shortcite{Zhou:2013} (top) and our approach (bottom) on the \textit{stonewall} example from their 3D Scene Dataset. This example consists of $2,700$ frames.}\label{fig:zhou13}
\end{figure}

\subsubsection{Reconstruction accuracy}
To evaluate the reconstruction ac\-curacy, we compared the results of \textit{Redwood}~\cite{choi2015robust}, \textit{ElasticFusion}~\cite{Whelan2016ElasticFusionRD}, \textit{BundleFusion}~\cite{Dai2017}, \textit{UncertaintyAware}~\cite{Cao2018}, and ours to the 3D models acquired by a commercial hand-held structure light scanner, Artec Eva, with the precision of $0.1mm$. The model obtained from this structure light scanner is referred to as ground truth. To evaluate the relevant scenery, we manually removed the background of the obtained model from each method. Each model is aligned to the ground truth mesh and the distance error is computed and visualized as a color map (see Fig.~\ref{fig:comparison_reconstruction_accuracy}). As can be observed, all these methods were able to produce consistent 3D models and our results have the smallest errors while preserving more geometric details than the other methods. The errors were mainly sourced to camera tracking, which is prone to noises on the input RGB-D images. By using the on-the-fly HRBF surface estimation together with the weighted registration strategy (i.e., curvature, confidence, and photometric), our system is more robust in camera tracking, therefore, yielding the highest precision among all these systems.

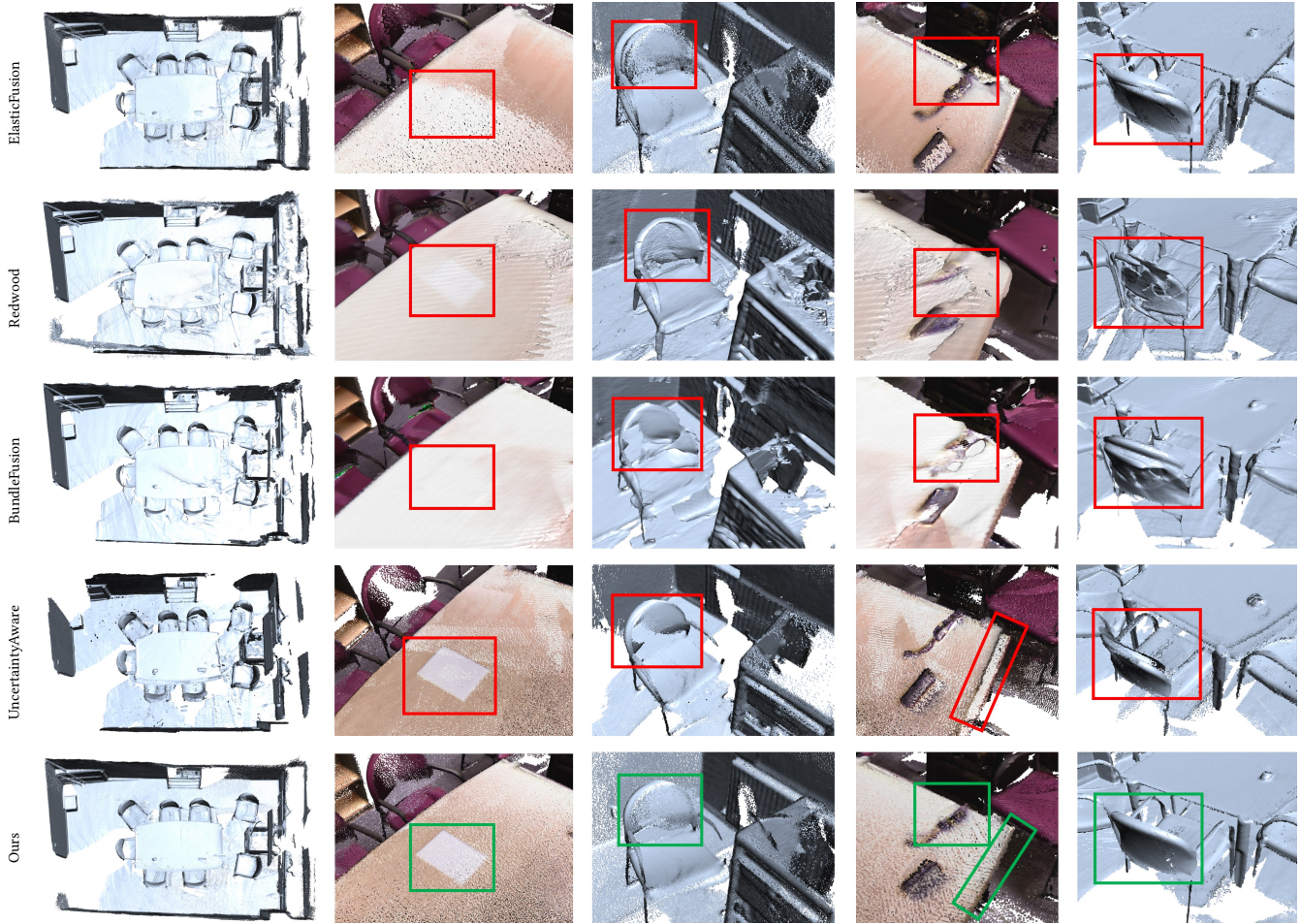
\begin{figure*}[htbp]
	\setlength{\unitlength}{0.1\textwidth}
	\begin{picture}(10.0, 7.2)
	\put(0.35,0.15){\includegraphics[width=0.96\linewidth]{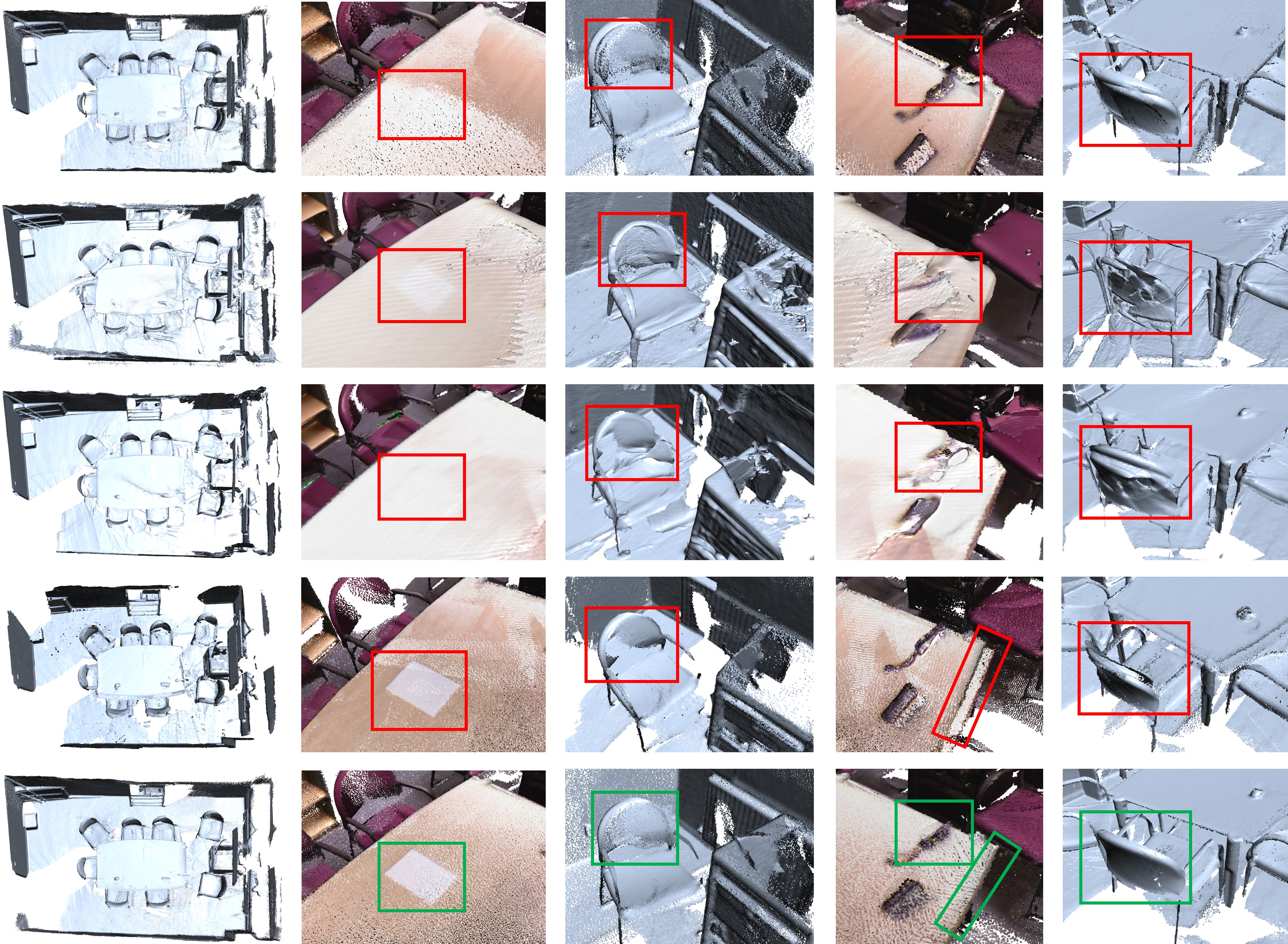}}
	\put(0.1,6.1){\scriptsize \rotatebox{90}{ElasticFusion}}
	\put(0.1,4.7){\scriptsize \rotatebox{90}{Redwood}}
	\put(0.1,3.2){\scriptsize \rotatebox{90}{BundleFusion}}
	\put(0.1,1.7){\scriptsize \rotatebox{90}{UncertaintyAware}}
	\put(0.1,0.6){\scriptsize \rotatebox{90}{Ours}}
	\end{picture}
	\vspace{-20pt}
	\caption{Comparison with state-of-the-art RGB-D reconstruction systems, i.e., \textit{ElasticFusion}~\cite{Whelan2016ElasticFusionRD},  \textit{Redwood}~\cite{choi2015robust}, \textit{BundleFusion}~\cite{Dai2017}, and UncertaintyAware~\cite{Cao2018} on a sequence of $6,114$ RGB-D images captured in a conference room by a complex camera trajectory that consists of many local loops (see Fig.~\ref{fig:confRoom_localloops}).}\label{fig:Redwood_ElasticFusion_BundleFusion_indoor_scene}
\end{figure*}
\begin{figure}
\includegraphics[width=\linewidth]{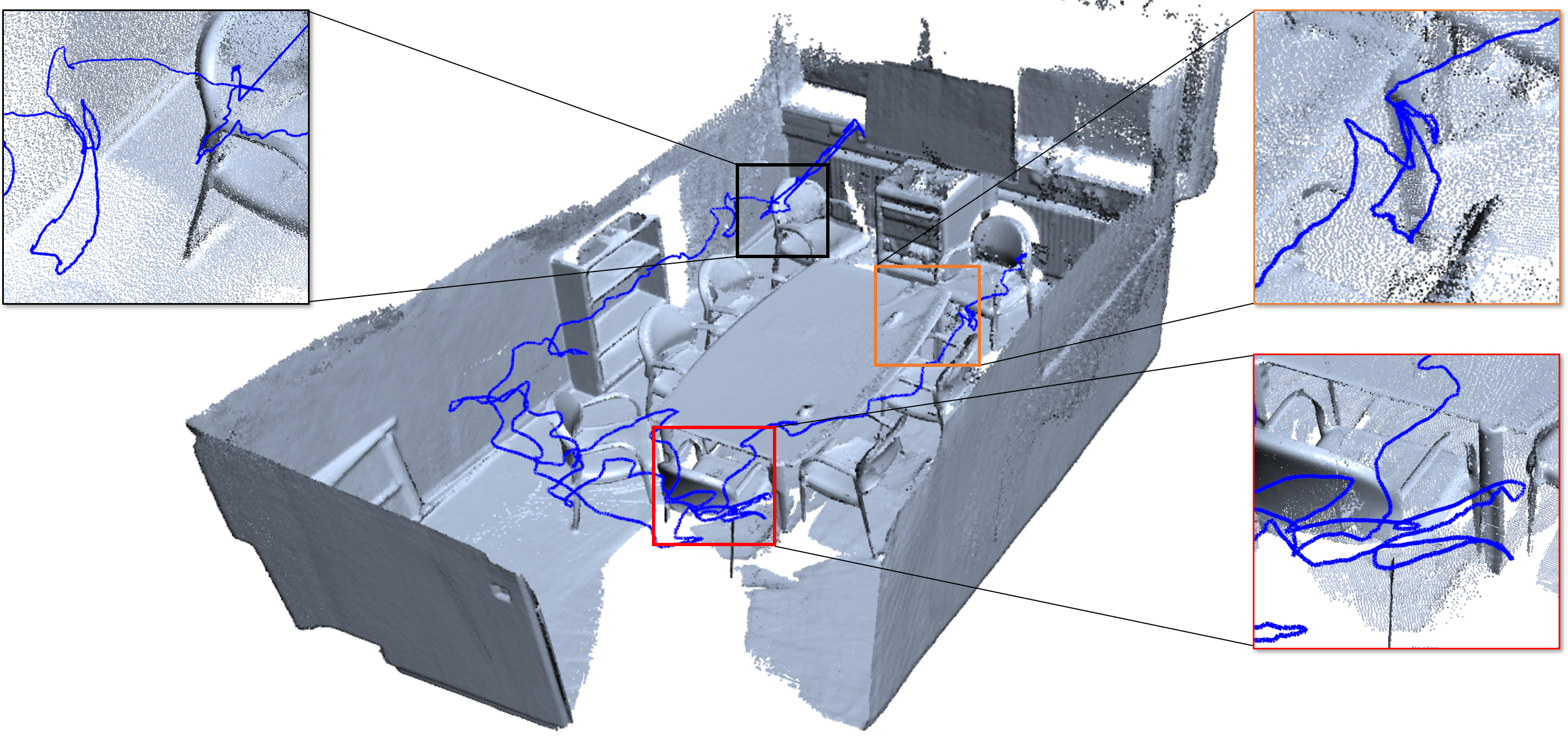}\\
\vspace{-8pt}
\caption{The complex camera trajectory for the conference room example shown in Fig.~\ref{fig:Redwood_ElasticFusion_BundleFusion_indoor_scene} with $6,114$ frames.}\label{fig:confRoom_localloops}
\end{figure}

We also evaluate the surface reconstruction accuracy in terms of average point-to-surface distances on the~\textit{living room kr0-kr3} models from the \textit{ICL-NUIM benchmark}~\cite{handa:etal:ICRA2014}. Our method is compared with a variety of existing approaches and the results are summarized in Fig.~\ref{fig:ICL-NUIM_surface_reconstruction_error}. It is easy to find that our method can achieve better (or comparable) results in terms of reconstruction accuracy. Again this is benefited from the robust HRBF on-the-fly surface estimation presented in this paper.

\begin{figure*}[htbp]
	\setlength{\unitlength}{0.1\textwidth}
	\begin{picture}(10,3.1)
	\put(0.35,0.04){\includegraphics[width=0.95\textwidth]{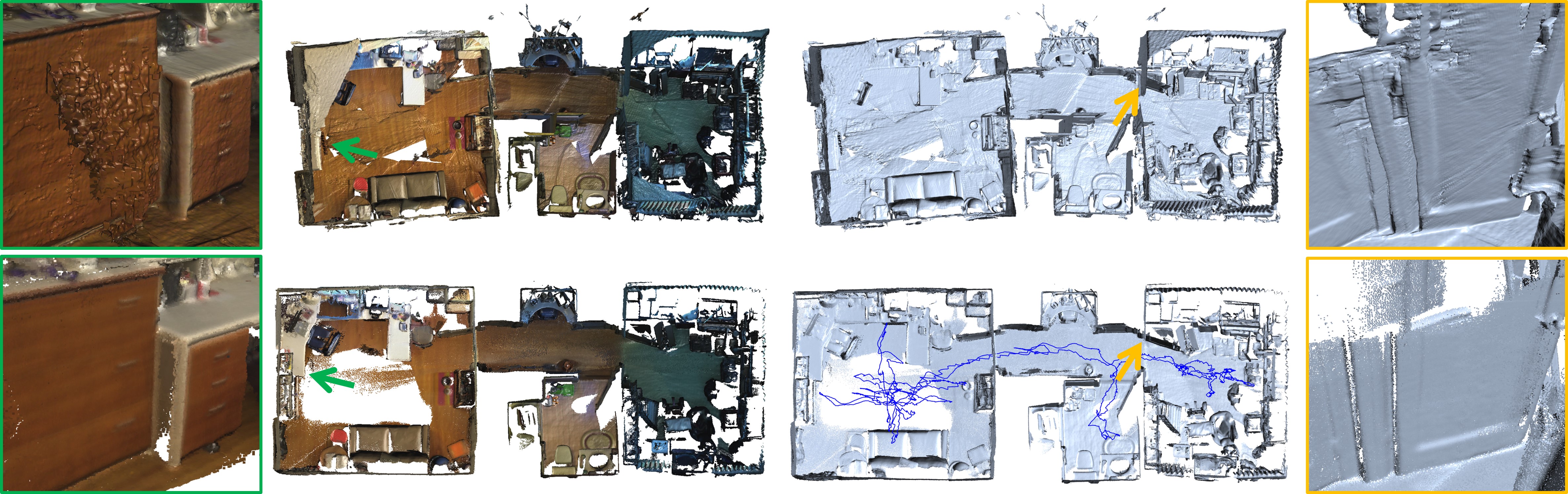}}
	\put(0.18, 0.6){\scriptsize \rotatebox{90}{Ours}}
	\put(0.18, 1.9){\scriptsize \rotatebox{90}{BundleFusion}}
	\end{picture}
\caption{Comparison of our reconstruction result (bottom row) with BundleFusion~\cite{Dai2017} (top row) on "scene0054\_00", a sequence of $6,629$ RGB-D frames captured by a structure sensor, from ScanNet~\cite{ScanNetDai2017}. Closer inspections (green and yellow boxes) are presented to show reconstruction details of each method. The camera trajectory is visualized in blue color. Our approach maintains only global model points with confidence values larger than a threshold (see Section~\ref{secDepthFusion}) similar to~\cite{Cao2018, Whelan2016ElasticFusionRD, Keller2013}, which causes some missing data on the floor.
	}\label{fig:Comparison_ScanNet_SequenceReconstruction}
\end{figure*}


\subsubsection{Scalability} 
With the robustness in camera tracking and surface prediction, our method can reconstruct large scenes. In addition to the two scenes already shown in Fig.~\ref{figTeaser}, we tested our approach on the \textit{stonewall} models from the 3D Scene dataset (see Fig.~\ref{fig:zhou13}). Comparing their method with the offline global optimizer \cite{Zhou:2013}, we can observe a significant reduction of camera tracking drift on our result on this example with $2,700$ frames. 

As shown in Figs.~\ref{fig:Redwood_ElasticFusion_BundleFusion_indoor_scene} and \ref{fig:confRoom_localloops}, we captured a sequence of $6,114$ RGB-D images in a conference room by a \textit{Microsoft Kinect v1} camera with a complex camera trajectory. The trajectory contains many local loops. When comparing with other state-of-the-art reconstruction systems including \textit{Redwood}~\cite{zhou2015}, \textit{ElasticFusion}~\cite{Whelan2016ElasticFusionRD}, \textit{BundleFusion}~\cite{Dai2017}), and UncertaintyAware~\cite{Cao2018}, all the other four methods suffer from camera tracking drift (especially in the regions with local loops on the trajectory) and perform poorly in recovering surface details -- see the `double-layers' of chairs (fourth column) and tables (fifth column) shown in the zoom-views. Our robust surface estimation by using on-the-fly HRBF implicits can effectively reduce the error in camera tracking drift thus can generate more consistent 3D reconstruction. 

\begin{figure*}[!tb]
 \centering
 \begin{subfigure}{0.49\linewidth}
  \centering
  \captionsetup{justification=centering}
  \includegraphics[width=\linewidth]{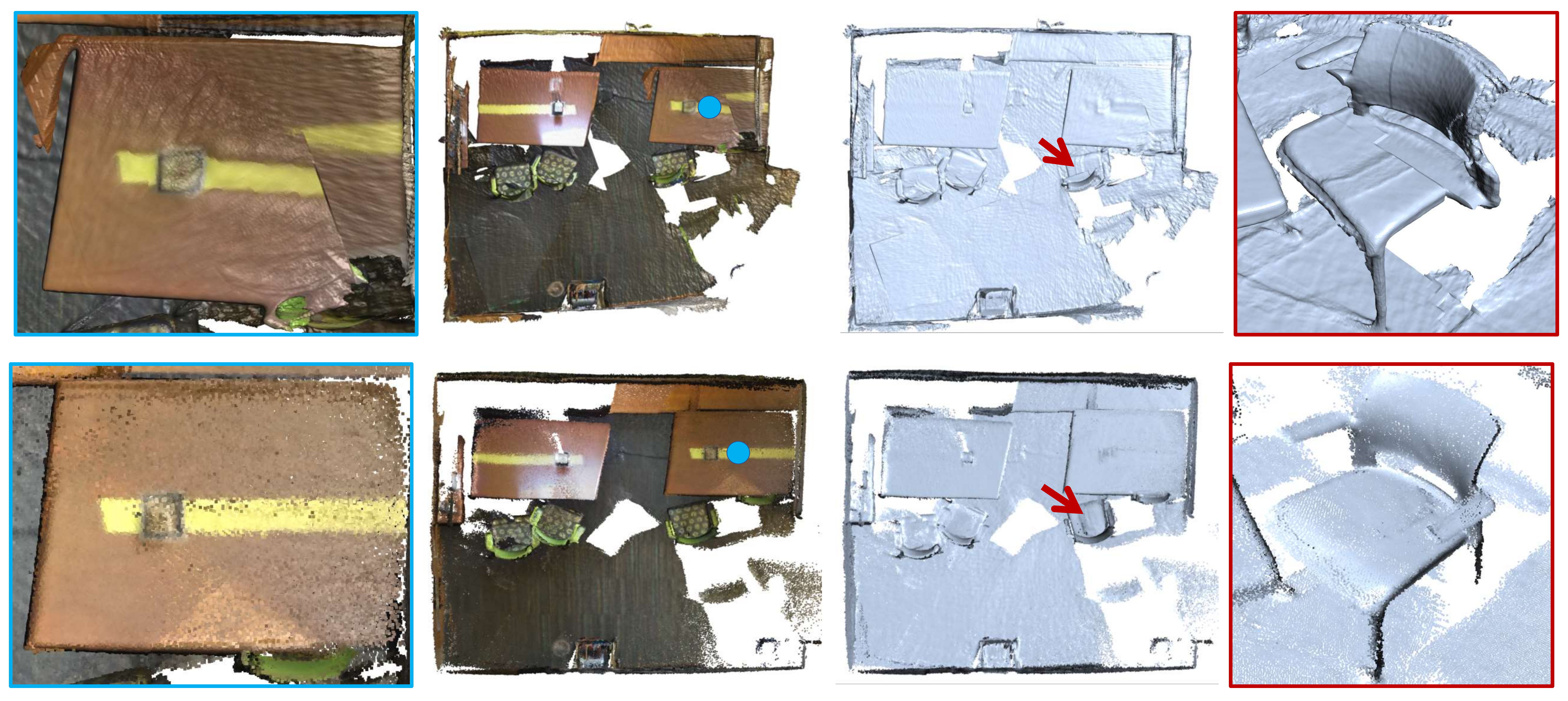}
  \caption{scene0005\_00}
  \label{fig:d1_m0_np0_po0}
 \end{subfigure}
 \hfill
 \begin{subfigure}{0.49\linewidth}
  \centering
  \captionsetup{justification=centering}
  \includegraphics[width=\linewidth]{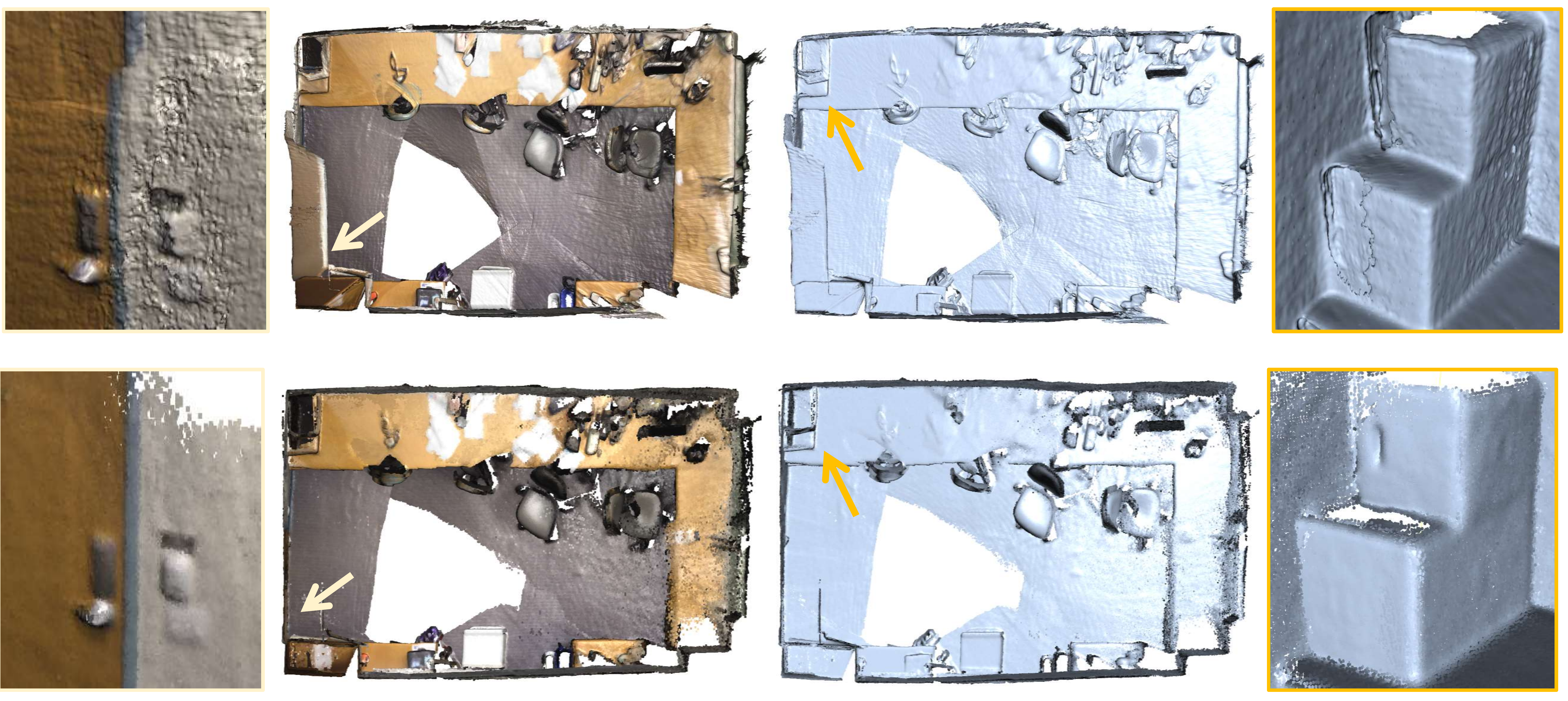}
  \caption{scene0098\_00}
  \label{fig:d2_m0_np0_po0}
 \end{subfigure}
 \begin{subfigure}{0.49\linewidth}
  \centering
  \captionsetup{justification=centering}
  \includegraphics[width=\linewidth]{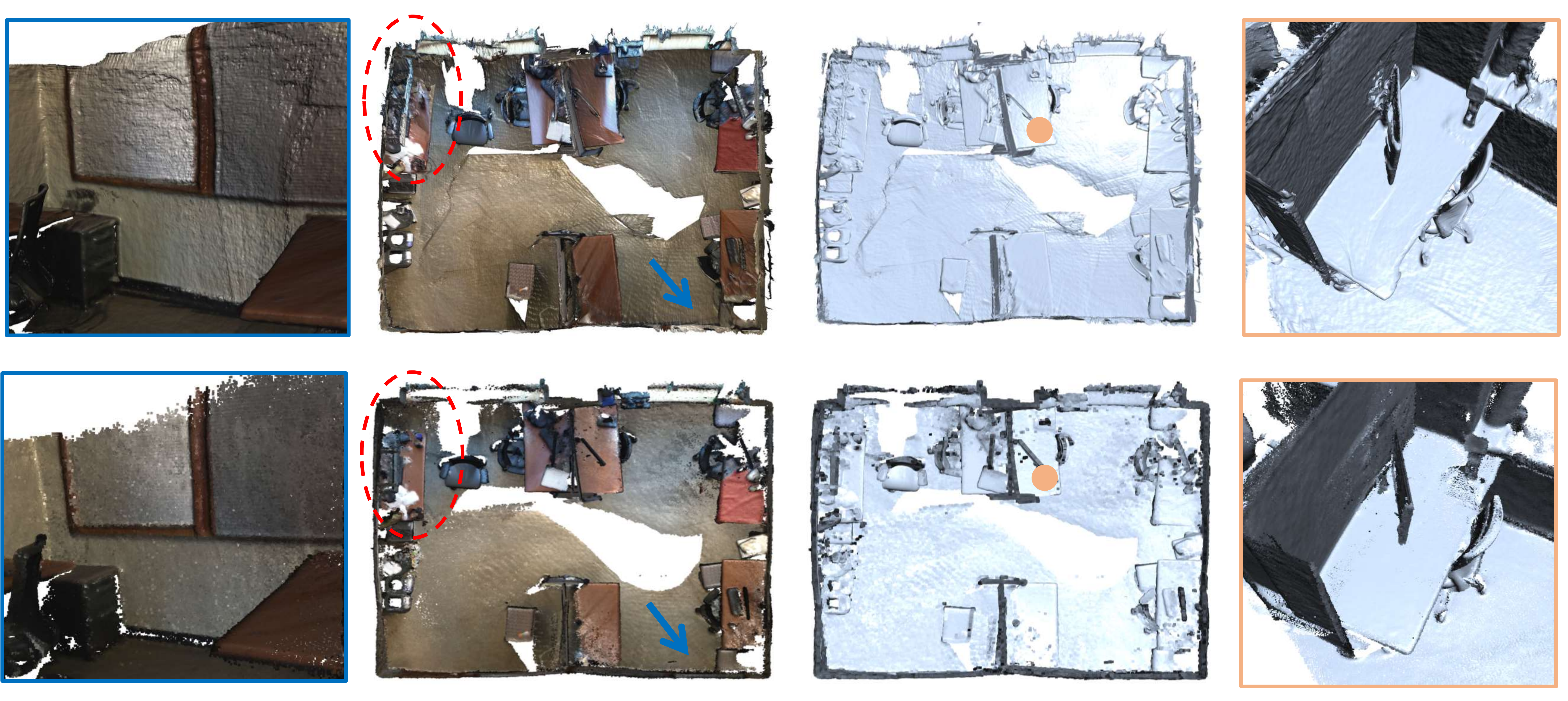}
  \caption{scene0142\_00}
  \label{fig:m1_np0_po0}
 \end{subfigure}
 \hfill
 \begin{subfigure}{0.49\linewidth}
  \centering
  \captionsetup{justification=centering}
  \includegraphics[width=\linewidth]{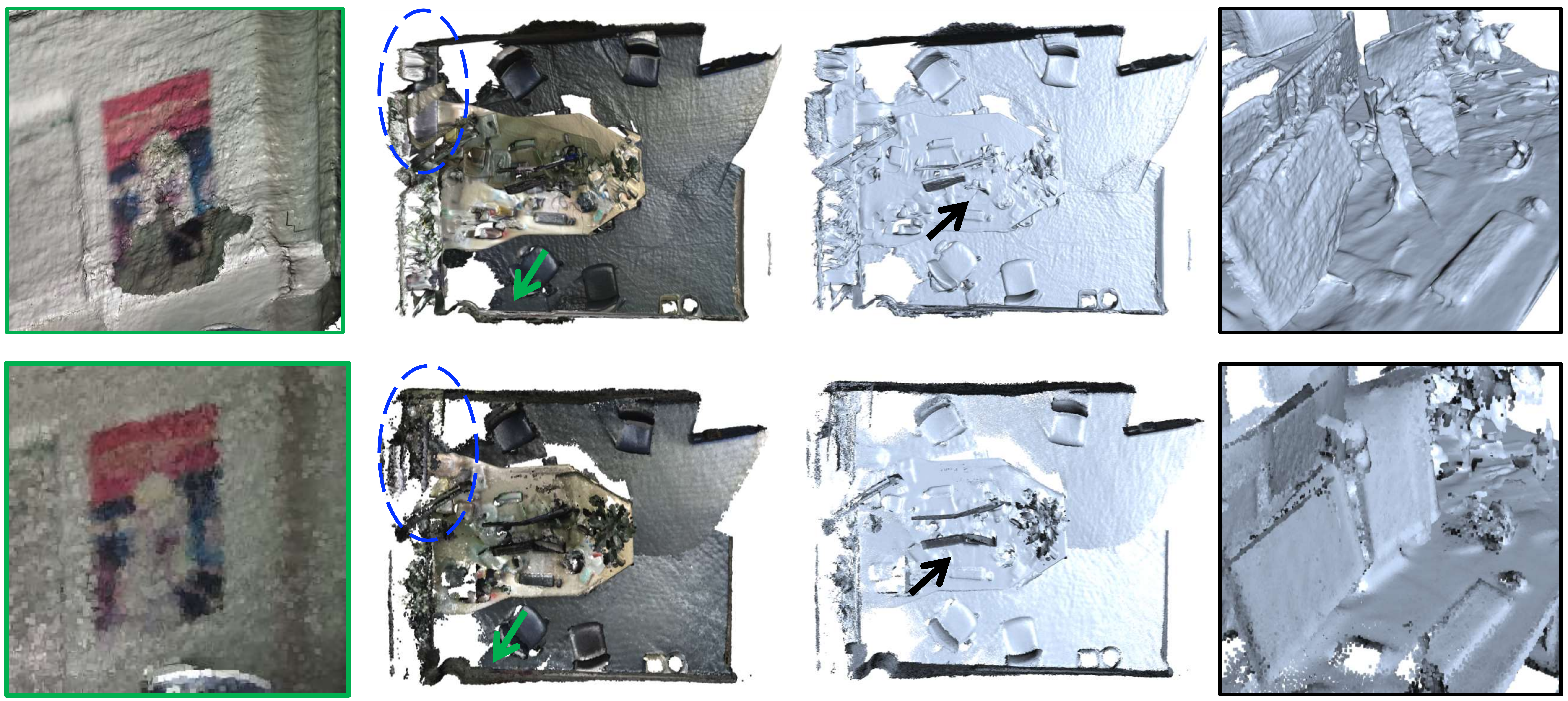}
  \caption{scene0017\_00}
  \label{fig:m0_np1_po0}
 \end{subfigure}
 \caption{More comparison between the reconstruction results of BundleFusion~\cite{Dai2017} (top) and ours (bottom) on the other four sequences from the ScanNet dataset~\cite{ScanNetDai2017}, which contain $1159$, $1285$, $2434$, and $1490$ RGB-D frames, respectively. Structural distortions are marked in colored circle, and closer zoom-views are also presented to show the details of 3D reconstruction.}
 \label{fig:Comparison_ScanNet_SequenceReconstruction2}
\end{figure*}

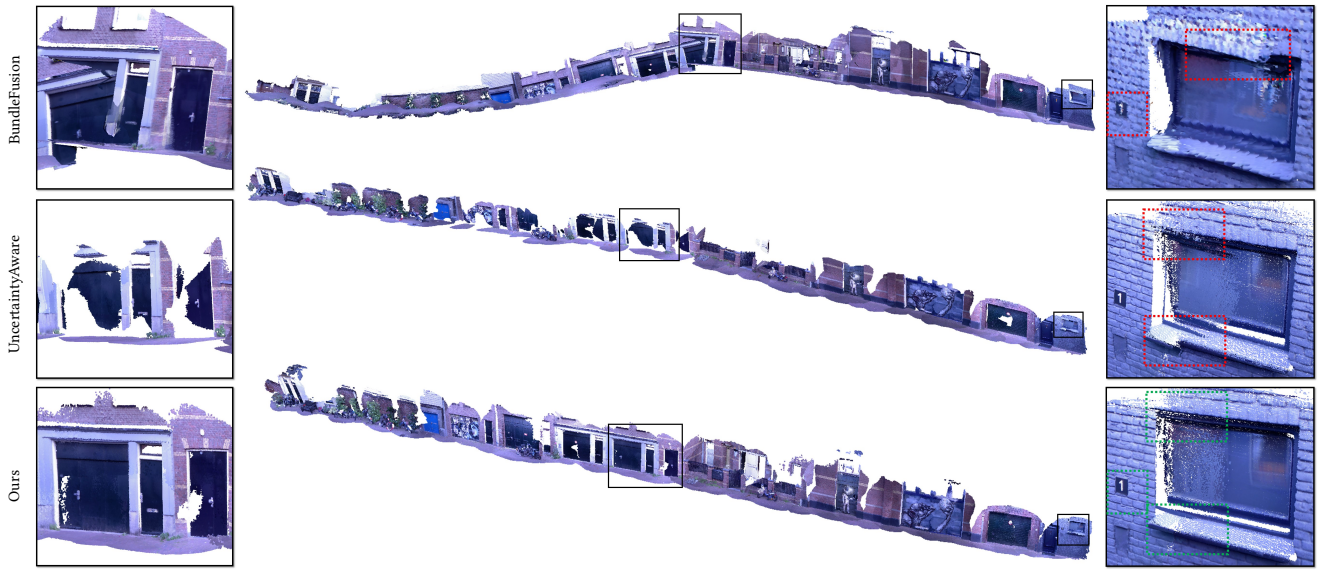
\begin{figure*}
	\setlength{\unitlength}{0.1\textwidth}
	\begin{picture}(10,4.45)
	\put(0.35,0.1){\includegraphics[width=0.95\textwidth]{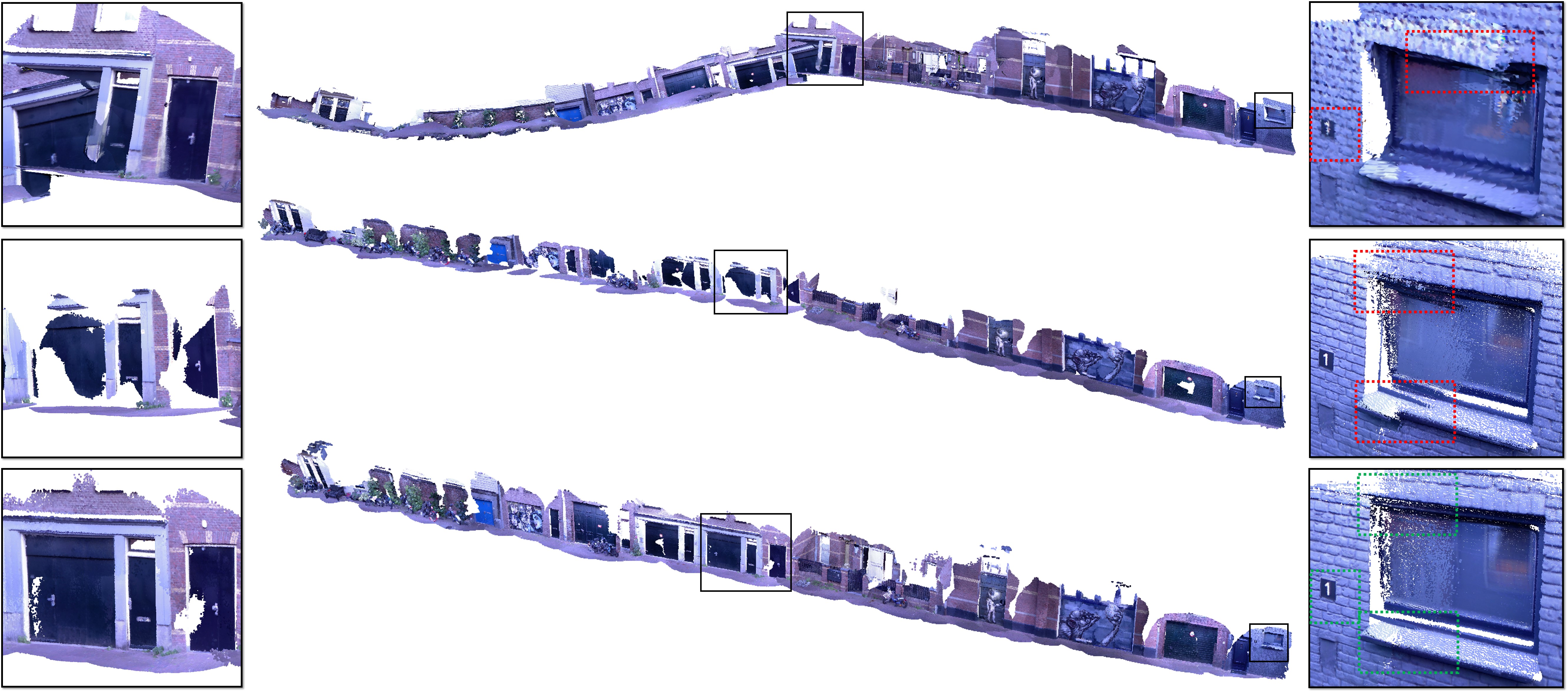}}
    \put(0.15, 0.60){\scriptsize \rotatebox{90}{Ours}}
	\put(0.15, 1.70){\scriptsize \rotatebox{90}{UncertaintyAware}}
	\put(0.15, 3.25){\scriptsize \rotatebox{90}{BundleFusion}}
	\end{picture}
	\vspace{-15pt}
	\caption{Comparison of our reconstruction result (bottom row) with state-of-the-art RGB-D reconstruction systems, i.e., \textit{BundleFusion}~\cite{Dai2017} (top row) and \textit{UncertaintyAware}~\cite{Cao2018} (center row) on a sequence of $14,163$ RGB-D frames captured on an urban street with a long trajectory. Closer inspections (black boxes) are presented to show the reconstructed details of each method.}\label{fig:evaluationOutdoorScene}
\end{figure*}
\begin{figure*}
	\setlength{\unitlength}{0.1\textwidth}
	\begin{picture}(10,5.6)
	\put(0.35,0.1){\includegraphics[width=0.95\textwidth]{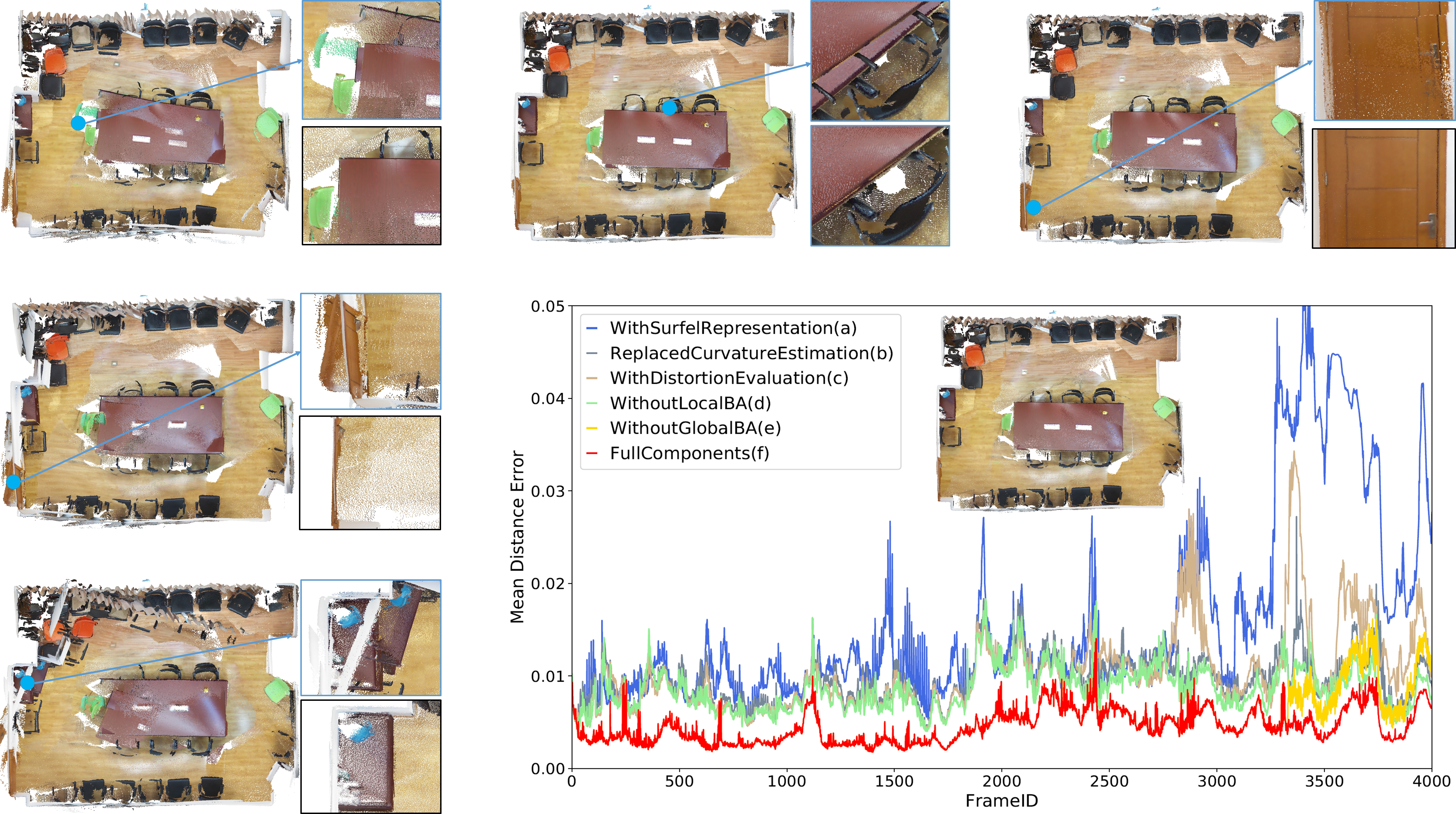}}
	\put(1.4, 3.65){\scriptsize (a)}
	\put(4.7, 3.65){\scriptsize (b)}
	\put(8.05, 3.65){\scriptsize (c)}
	\put(1.4, 1.8){\scriptsize (d)}
	\put(1.4, -0.05){\scriptsize (e)}
	\put(7.2, 1.9){\scriptsize (f)}
	\put(6.85, -0.05){\scriptsize (g)}
	\end{picture}
	\caption{Evaluation of our system with different options for each single component on a sequence of 4,080 RGB-D images captured in another meeting room (different from Fig.~\ref{fig:confRoom_localloops}). (a) Replacing HRBF implicits by the surfel-based representation. (b) Changing HRBF-based curvature to the curvature estimation method presented in~\cite{lefloch2017}. (c) Using the camera-distortion based confidence evaluation method in~\cite{Keller2013}. (d) Removing local BA. (e) Removing global BA. (f) Our reconstruction result with full components. (g) Mean distance error of all validated vertex pairs (see in Section~\ref{subsecCorrespondSearch}) for each frame pair. The close-up views in (a)-(e) show the artifacts obtained by changing one component (top) and our corresponding reconstruction (bottom) using all algorithm components.}\label{fig:evaluationSingleComponent}
\end{figure*}

We also conducted experiments on the ScanNet dataset~\cite{ScanNetDai2017}. Among its 1,500 scan sequences, we randomly selected 200 sequences to reconstruct 3D scenes and compared our results with those from BundleFusion~\cite{Dai2017}. It is found that similar results are generated by both methods on most of the sequences especially on those for simple scenes. Better reconstruction results can be found on 5 sequences with complex trajectories. For example, in the scene shown in Fig.~\ref{fig:Comparison_ScanNet_SequenceReconstruction}, the structural distortion is significantly reduced by our method. Similar improvement can also be found on the other four scenes as shown in Fig.~\ref{fig:Comparison_ScanNet_SequenceReconstruction2}.

Moreover, we captured a sequence of $14,163$ RGB-D frames with a quite long trajectory on an urban street using a~\textit{Microsoft Kinect v1}. We compare the reconstruction results with \textit{BundleFusion}~\cite{Dai2017} and \textit{UncertaintyAware}~\cite{Cao2018} in Fig.~\ref{fig:evaluationOutdoorScene}. We can observe that the result of \textit{BundleFusion}~\cite{Dai2017} breaks (see the zoom-view on the top left) and fails to generate a globally consistent 3D model. \textit{UncertaintyAware} can obtain a more consistent result but still suffers from camera tracking drift, which leads to artifacts in the reconstruction (see the zoom-view on the right of the second row). In contrast, our system can produce a globally consistent 3D model. It is also worthy to note that surfel-based representation has the advantage to preserve geometric details. This can be observed from the number plate `1' in the right zoom-views, where the result obtained from the volumetric representation of \textit{BundleFusion} is not as clear as \textit{UncertaintyAware} and ours.

\subsubsection{Ablation Study}
We further conducted an ablation study to evaluate the effectiveness of each single algorithm component of our system by replacing it with another option used in others' work, where the study is taken on a sequence of $4,080$ RGB-D images captured in a meeting room (Fig.~\ref{fig:evaluationSingleComponent}). For quantitative analysis, we also plot the mean distance errors of all validated vertex pairs for each frame pair to indicate the quality of the registration as shown in Fig.~\ref{fig:evaluationSingleComponent}(f). Due to the high sensitivity to noise, the surfel-based representation led to a dramatic increase in mean distance error (see Fig.~\ref{fig:evaluationSingleComponent}(f)) and unsatisfactory reconstruction (see the close-up view shown in Fig.~\ref{fig:evaluationSingleComponent}(a)). In the second test, we replace our HRBF-based curvature estimation with the method presented in \cite{lefloch2017}. Although the global consistent 3D model can be obtained -- thanks to global techniques of local and global BA, noises induced by the black surfaces (i.e., chairs) can lead to artifacts in intra-submap level as shown in Fig.~\ref{fig:evaluationSingleComponent}(b). In Fig.~\ref{fig:evaluationSingleComponent}(c), we utilize the camera-distortion based evaluation method~\cite{Keller2013} for confidence map. As a result, the accumulated noise in the global model leads to unstable registration and imperfect reconstruction. The last two tests are conducted to evaluate the importance of local BA (Fig.~\ref{fig:evaluationSingleComponent}(d)) and global BA (Fig.~\ref{fig:evaluationSingleComponent}(e)) in our pipeline of reconstruction, where local BA helps to recover the artifacts between submaps and global BA helps to generate globally consistent models.

\begin{figure}
\centering
\includegraphics[width=\linewidth]{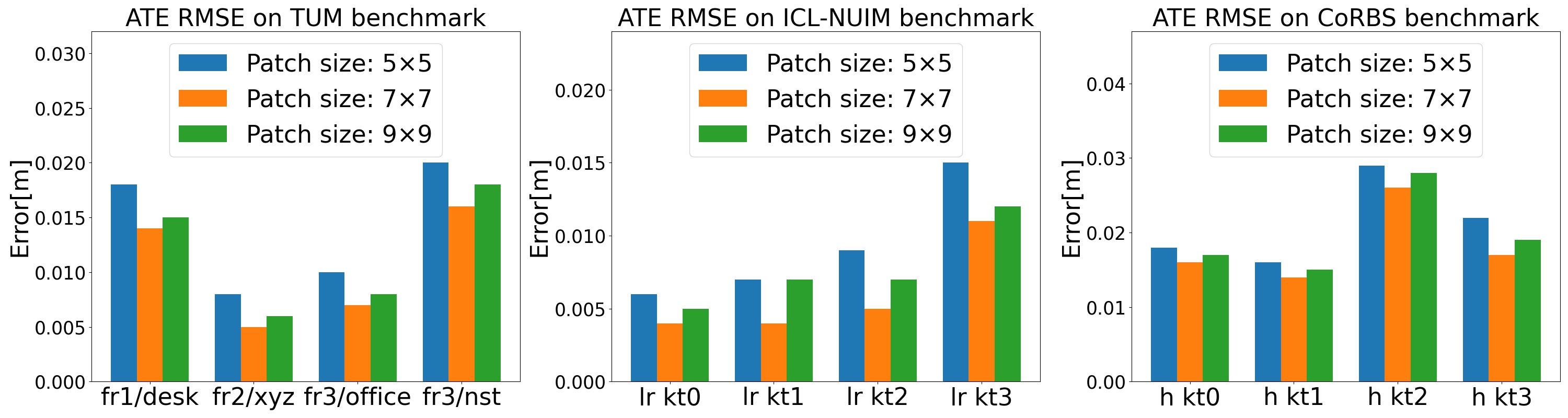}
\caption{The evaluation of using different support sizes on the accuracy of camera tracking on different datasets, i.e., \textit{TUM benchmark}, \textit{ICL-NUIM benchmark}, and \textit{CoRBS benchmark}. We select different sizes of window patches (as described in Section~\ref{subsecSurfaceEvaluation}): $5 \times 5$,  $7 \times 7$, and $9 \times 9$ for the evaluation.}\label{fig:support_size}
\end{figure}

\subsubsection{Parameter discussion.} 
As a key parameter of our system, the support size influences the accuracy of registration. We select different sizes of window patches (Section~\ref{subsecSurfaceEvaluation}) as $5 \times 5$,  $7 \times 7$ (default), and $9 \times 9$ for the experiment. The evaluation is conducted on three different datasets, i.e., \textit{TUM benchmark}, \textit{ICL-NUIM benchmark}, and \textit{CoRBS benchmark}, and the results are presented in Fig.~\ref{fig:support_size}. A larger support size leads to a smoother surface while a smaller support size can preserve more geometric details. Correspondingly, a patch size of $5 \times 5$ is suitable for the reconstruction of clean data but not robust enough to handle the noises induced by depth cameras. On the other hand, a patch size of $9 \times 9$ always leads to over-smoothing results. A patch size of $7 \times 7$ can achieve the best performance in our tests as shown in Fig.~\ref{fig:support_size}.

\subsubsection{Performance} To study the memory consumption of our approach, we recorded the GPU memory consumption for storing and managing the global model over $13$ sequences of RGB-D images that are captured. Comparison with the \textit{BundleFusion} system~\cite{Dai2017} is conducted to demonstrate the memory efficiency of our HRBF-based method -- see Table~\ref{table:memory consumption}. Note that \textit{BundleFusion} shares the same representation with \textit{VoxelHashing}~\cite{NieBner2013}, which exploits sparsity by applying a hash-based structure to the volumetric representation. The memory consumption of \textit{BundleFusion} by using two different voxel sizes is reported. When $4$mm is used for the voxel size -- being able to capture more geometric details, the \textit{BundleFusion} system failed to add depth maps during reconstruction due to the large memory requirement. Our system has a significantly smaller memory footprint compared to the volumetric representation based approaches and therefore is more suitable for reconstructing large scenes.

\begin{table}[htbp]
	\caption{Statistic of memory consumption for reconstruction (unit: MB)}
	\label{table:memory consumption}\vspace{-8pt}\centering\small
	\begin{tabular}{r|c|r|r|r|r}
	\hline
	Model Name  & Fig. & \#Frames & \multicolumn{2}{c|}{BundleFusion (GPU)} & Ours (GPU) \\ 
	\hline
	\hline
			   & & & \multicolumn{2}{c|}{Voxel Size}    &  \\ 
			   \cline{4-5}
		       &  &  & \hspace{5pt} 10mm  & 4mm    &  \\ 
	\hline
	\textit{Fertility}  &  \ref{fig:Comparison_Redwood_BundleFusion_object_reconstruction}   & 1,301 &     153.0 &    740.9 &  15.0\\ 
	\hline
	\textit{Plant}	  &   \ref{fig:Comparison_Redwood_BundleFusion_object_reconstruction}   & 1,703 &     162.9 &    830.5 &  25.3\\ 
	\hline
	\textit{Human}	  &   \ref{fig:Comparison_Redwood_BundleFusion_object_reconstruction}   & 2,751 &     137.4 &    1309.1 &  44.9\\ 
	\hline
	\textit{Pillar}	  &   \ref{fig:Comparison_Redwood_BundleFusion_object_reconstruction}   & 1,987 &     198.2 &    1536.6 &  59.8\\ 
	\hline
	\textit{Car Frame}  &   \ref{fig:Comparison_Redwood_BundleFusion_object_reconstruction} 	& 3,694 &     126.8 &    1462.5 &  95.1\\ 
	\hline
	\textit{Faces}  &   \ref{fig:comparison_reconstruction_accuracy} 	& 579 &  59.0    &  443.0  &  21.1\\ 
	\hline
	\textit{Head}  &   \ref{fig:comparison_reconstruction_accuracy} 	& 1,663 &  167.6    &  1806.4   &  15.6\\ 
	\hline
	\textit{Upper Body} &   \ref{fig:comparison_reconstruction_accuracy} 	& 1,308 &   179.7   &  1951.6   &  22.3\\
	\hline
	\textit{Small Chair} &   \ref{fig:comparison_reconstruction_accuracy} 	& 1,693 &   107.5   &   1115.2  &  34.0\\ 
	\hline
	\textit{Conference Room} & \ref{fig:Redwood_ElasticFusion_BundleFusion_indoor_scene}	& 6,114 &     164.7 &    1240.3 &  98.1\\
	\hline
	\textit{Urban Street} & \ref{fig:evaluationOutdoorScene}	& 14,163 &  2205.3  &    - &  552.5\\
	\hline
	\textit{Library}	 &  \ref{figTeaser} & 16,128 &    1924.1 &  -  &  571.3\\ 
	\hline
	\textit{Study Platform} &  \ref{figTeaser}	& 10,930 &   1835.0 &  -  & 603.1\\
	\hline
	\end{tabular}
\end{table}

\begin{figure*}[htbp]
	\setlength{\unitlength}{0.1\textwidth}
	\begin{picture}(10,2.6)
	\put(0.35,0.1){\includegraphics[width=0.95\textwidth]{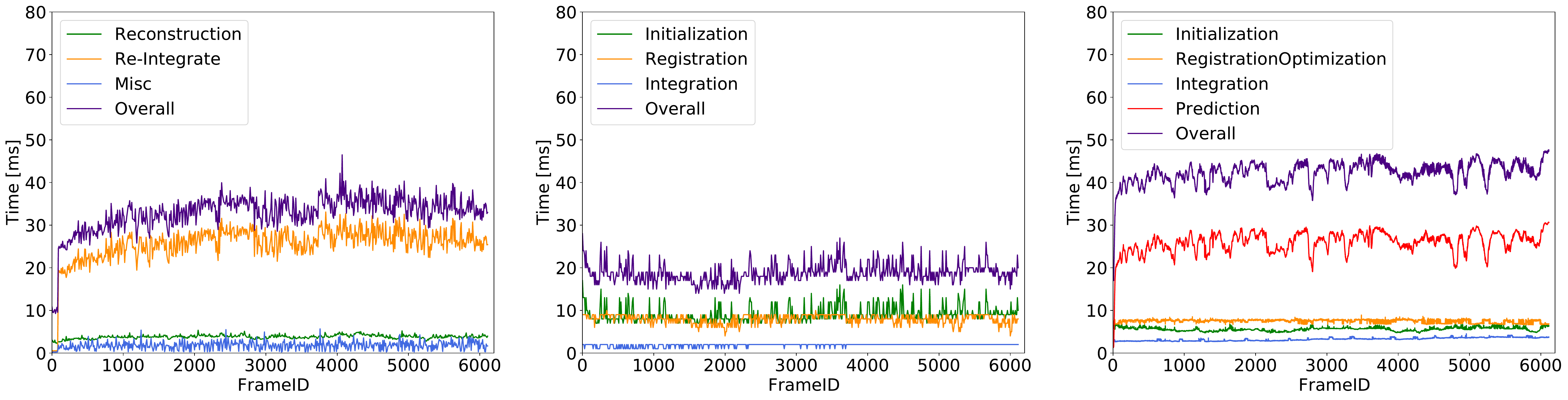}}
	\put(1.4, 0.0){\scriptsize BundleFusion~\cite{Dai2017}}
	\put(4.5, 0.0){\scriptsize UncertaintyAware~\cite{Cao2018}}
	\put(8.3, 0.0){\scriptsize Ours}
	\end{picture}
	\caption{Comparison of the computing time of our system (right) with \textit{BundleFusion}~\cite{Dai2017} (left) and \textit{UncertaintyAware}~\cite{Cao2018} (center) on each frame of the~\textit{Meeting Room} sequence (Fig.~\ref{fig:Redwood_ElasticFusion_BundleFusion_indoor_scene}). The processing time for each main component of the three systems is also plotted. Note that, two GPUs are used for~\textit{BundleFusion} as suggested in \cite{Dai2017} and the time is reported here according to the main GPU. Differently, only one GPU is employed for \textit{UncertaintyAware} and our pipeline.
	}\label{fig:running_time}
\end{figure*}

We report the computing time used by each component of our system in Fig.~\ref{fig:running_time} for all RGB-D frames throughout the sequence of~\textit{Meeting Room} (Fig.~\ref{fig:Redwood_ElasticFusion_BundleFusion_indoor_scene}). The efficiency of different components of the computational pipeline has been analyzed. In general, our system can achieve an average processing time of 42ms per frame, which indicates a near real-time performance (i.e., approximately 24Hz). Among all components of our system, the HRBF-based prediction takes over half of the processing time (25ms). For comparisons, we also plot the processing times of \textit{BundleFusion}~\cite{Dai2017} (left) and \textit{UncertaintyAware}~\cite{Cao2018} (center) in Fig.~\ref{fig:running_time}. 


%
\section{Conclusion and Future Work}
\label{secConclusion}
We have presented the HRBF-Fusion as a new method using on-the-fly HRBF implicits for 3D reconstruction from RGB-D images. Our system is not only able to reconstruct objects with high fidelity but also scalable to large scenes after incorporating submap-based local and global optimization strategies. The robustness of our HRBF-Fusion is mainly due to the robust curvature estimation based on the HRBF implicits, which can significantly reduce the drift in camera tracking. Moreover, our reconstruction-indicated surface evaluation method exploits the uncertainty of the measurement in the input depth maps and further improves the accuracy in both the camera tracking step and the finally reconstructed models. The surfel representation using on-the-fly HRBF implicits has a low memory footprint and is suitable for reconstructing large scenes. 

The proposed system can reconstruct long-range scanning with submap level local and global optimization. However, camera tracking failure may still happen between intra-submaps for featureless regions (e.g., white planar walls). This is a common problem in all existing RGB-D reconstruction systems. A proactive reconstruction method by using robotic systems is planned to be investigated in our future work. Furthermore, it is also interesting to incorporate geometric primitives or structural regularities (i.e., parallelism or orthogonality) to improve the robustness of the hierarchical optimization for long-range scanning.

\begin{acks}
This project was financially supported in part by the National Key Research and Development Program of China (No. 2019YFB1707501), National Natural Science Foundation of China (No.~61772267), and Natural Science Foundation of Jiangsu Province (No. BK20190016). Yabin Xu was a visiting PhD student supervised by L. Nan and C.C.L. Wang at TU Delft, and he was also partially supported by the China Scholarship Council and the Faculty of Industrial Design Engineering, TU Delft. 
\end{acks}

\balance
\bibliographystyle{ACM-Reference-Format}
\bibliography{HRBF_Fusion_TOG}

\appendix

\vfill

\end{document}